\documentclass[11pt]{article}

\usepackage[final]{acl}

\usepackage{times}
\usepackage{latexsym}
\usepackage{booktabs}
\usepackage{multirow}
\usepackage{colortbl}
\usepackage{algorithm}
\usepackage{algpseudocode}
\usepackage{amsmath}

\usepackage[T1]{fontenc}
\usepackage[utf8]{inputenc}
\usepackage{enumitem}
\usepackage{hyperref}

\usepackage{microtype}

\usepackage{inconsolata}

\usepackage{graphicx}
\usepackage{subcaption}

\title{Efficient Test-Time Scaling via Temporal Reasoning Aggregation}

\author{
Jiakun Li\textsuperscript{1},
Xingwei He\textsuperscript{3},
Kefan Li\textsuperscript{1},
Hongzheng Chai\textsuperscript{1},
Hongyue Yu\textsuperscript{1},
Yuan Yuan\textsuperscript{1,2,3}\thanks{Corresponding author.} \\
\textsuperscript{1}School of Computer Science and Engineering, Beihang University \\
\textsuperscript{2}Qingdao Research Institute, Beihang University \\
\textsuperscript{3}Hangzhou Innovation Institute, Beihang University \\
\texttt{lijiakun25@buaa.edu.cn, hexingwei15@gmail.com, yuan21@buaa.edu.cn}
}

\begin{document}
\maketitle
\begin{abstract}

Test-time scaling improves the reasoning performance of large language models but often results in token-inefficient overthinking, where models continue reasoning beyond what is necessary for a correct answer. 
Existing dynamic early-exit methods typically rely on single-step confidence signals, which are often unreliable for detecting reasoning convergence in multi-step settings.
To mitigate this limitation, we propose TRACE, a training-free framework for efficient test-time scaling that determines when to terminate reasoning based on temporal aggregation of multi-step evidence rather than instantaneous signals.
TRACE detects reasoning convergence over time by aggregating two complementary signals across recent reasoning steps: answer consistency, capturing the persistence of predicted answers, and confidence trajectory, modeling the temporal evolution of model confidence. 
Benefiting from these two factors, TRACE can accurately determine whether the reasoning process has converged, thereby promptly halting inference and effectively avoiding redundant reasoning steps.
Extensive experiments on multiple challenging benchmarks show that TRACE reduces reasoning token usage by 25–30\% on average while maintaining accuracy within 1–2\% of full-length reasoning, outperforming existing dynamic reasoning methods. 
\footnote{To facilitate reproducibility, our code is available at \url{https://github.com/qianfantianyuzhouzhou/TRACE}.}

\end{abstract}

\section{Introduction}

The emergence of large reasoning models \citep{xu2025largereasoningmodelssurvey, yang2025qwen3technicalreport} has brought remarkable progress in solving complex tasks such as mathematical problem solving \citep{guan2025rstar} and code generation \citep{li2025cocoevo}. These approaches leverage the test-time scaling law \citep{snell2024scalingllmtesttimecompute} by increasing inference-time compute in two complementary ways: (i) extending the length of generated reasoning traces via chain-of-thought prompting \citep{wei2023chainofthoughtpromptingelicitsreasoning}, and (ii) expanding the search over alternative reasoning paths through self-consistency \citep{wang2023selfconsistencyimproveschainthought} or Tree of Thought \citep{yao2023tree}.
Test-time scaling has proven effective in enhancing the reasoning capability of Large Language Models (LLMs), and is now central to many state-of-the-art approaches.

However, test-time scaling tends to result in token-inefficient reasoning processes, thereby incurring unnecessary computational overhead or additional serving cost (e.g., API usage).
Specifically, LLMs may generate substantially longer chain-of-thought than necessary to reach a correct answer \citep{cuadron2025danger}, resulting in substantial token overuse with marginal accuracy gains—a phenomenon known as overthinking \citep{sui2025stop, chen2024not}.

\begin{figure}[t]
    \centering
    \includegraphics[width=\linewidth]{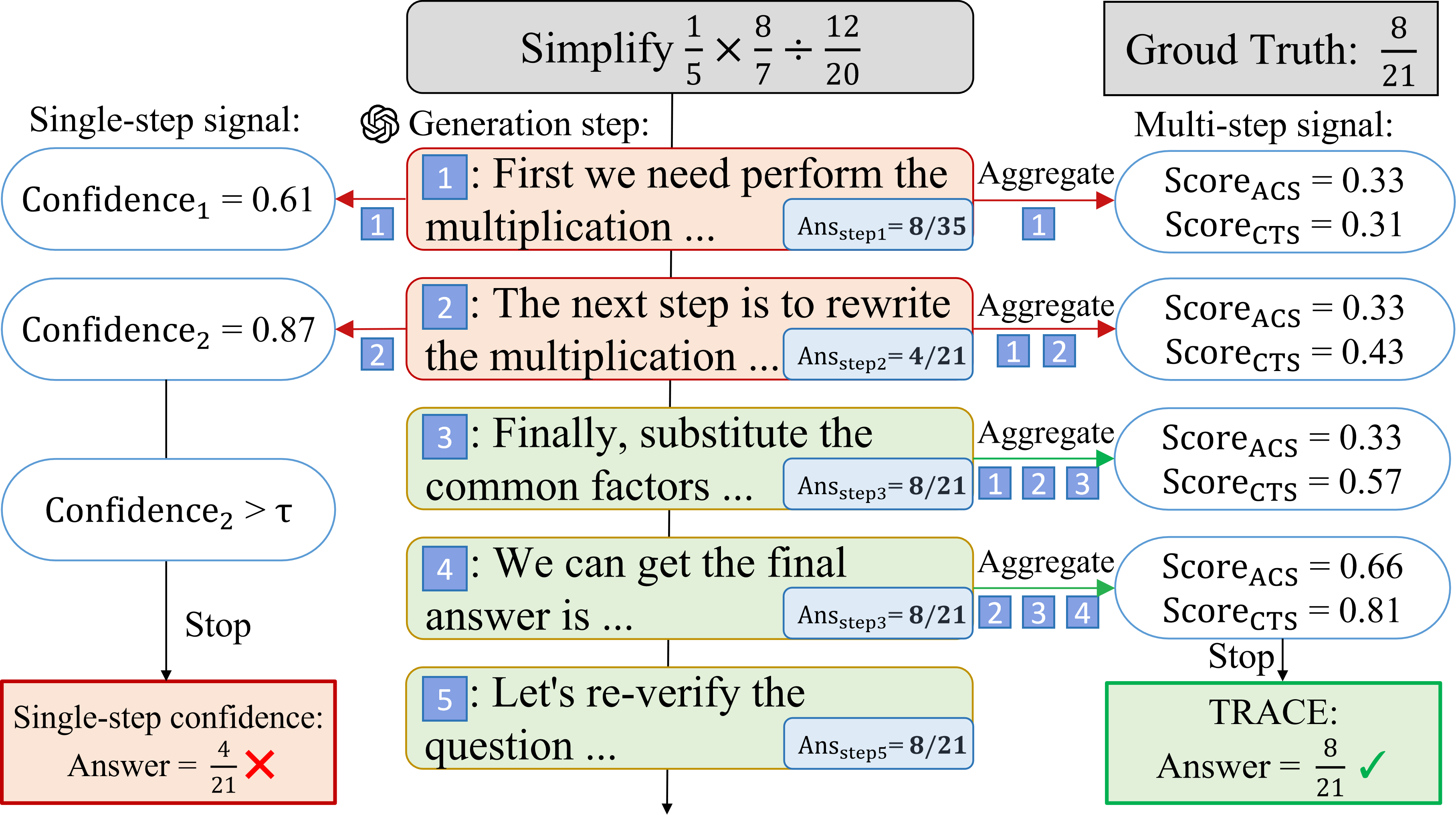}
    \vspace{-1.5em}
    \caption{
    \textit{Single-step confidence-based early-exit} prematurely stops at a high-confidence but incorrect intermediate prediction, while TRACE avoids false convergence and reaches the correct answer. The green box indicates the correct answer, and the red box indicates an incorrect answer.
    }
    \vspace{-1.55em}
    \label{fig:intro}
\end{figure}

Prior work has proposed both training-time and inference-time approaches to mitigate overthinking \citep{yang2025towards, lu2025retro, liu2025answerconvergencesignalearly}.
Training-time approaches explicitly reduce reasoning length by enforcing fixed token budgets \citep{han2025token} or by learning shorter reasoning trajectories via reinforcement learning \citep{arora2025training}.
In contrast, inference-time approaches mitigate overthinking via early-exit policies that terminate generation at inference time once sufficient evidence has been accumulated \citep{fu2024efficiently, huang2025efficient, yang2025dynamic}.
Such early-exit policies provide fine-grained control and can be applied to off-the-shelf models without additional training.

However, most existing dynamic early-exit methods rely primarily on single-step confidence signals. Recent studies \citep{xiong2023can, lacombe2025overreasoning_impairs_calibration} have revealed that such signals are unreliable during multi-step reasoning. Because reasoning convergence inherently requires stability across multiple steps, single-step confidence signals fail to indicate convergence as they reflect only per-step certainty rather than cross-step stability.
Consequently, relying solely on the unreliable single-step confidence signals forces LLMs into a dilemma between effectiveness and efficiency. Premature termination may halt reasoning too early, leading to incorrect outputs, whereas delayed termination prolongs computation unnecessarily, wasting resources; both cases ultimately undermine overall reasoning performance.
Figure~\ref{fig:intro} illustrates that the confidence calculated based on the single-step method is unreliable, as it assigns excessively high confidence to incorrect steps, thereby triggering the termination condition prematurely.

Motivated by this observation, we propose \textbf{Temporal Reasoning Aggregation for Convergent Exit (TRACE)}, a multi-step dynamic early-exit method that determines when to terminate reasoning based on the stability of model outputs across a window of recent steps. 
Since single-step confidence does not reliably indicate convergence in multi-step reasoning, TRACE provides more reliable signals of reasoning convergence by capturing multi-step answer consistency and the temporal evolution of confidence.
To be concrete, TRACE evaluates two complementary multi-step signals within a sliding window of recent reasoning steps. First, inspired by self-consistency \citep{wang2023selfconsistencyimproveschainthought}, we define the \textit{Answer Consistency Score (ACS)} to quantify cross-step agreement of induced answers. Furthermore, to mitigate the unreliable single-step signals, we introduce the \textit{Confidence Trajectory Score (CTS)} to track confidence evolution across steps.
By jointly considering answer consistency and confidence dynamics, TRACE achieves a better effectiveness--efficiency trade-off than single-step confidence-based methods.

In summary, this paper makes the following contributions:
\begin{itemize}[leftmargin=*, itemsep=3pt, parsep=0pt, topsep=4pt]
    \item We identify a risk that single-step confidence can be a misleading stopping signal, resulting in suboptimal early-exit decisions in reasoning.
    \item We propose a novel early-exit criterion that jointly considers the trajectory of model confidence over recent reasoning steps and the consistency of predicted answers, providing a more reliable signal of reasoning convergence than single-step confidence-based methods.
    \item We conduct extensive experiments on multiple reasoning benchmarks, proving that TRACE achieves a superior accuracy--efficiency trade-off. Compared to the strongest early-exit baselines, TRACE improves average accuracy by 2-4 points at comparable or lower token budgets. Meanwhile, relative to full-length reasoning, TRACE reduces reasoning tokens by 25--30\% on average while maintaining accuracy within 1--2\%.
\end{itemize}

\section{Evaluation of Single-Step Early-Exit}

\begin{figure}
    \centering
    \includegraphics[width=1\linewidth]{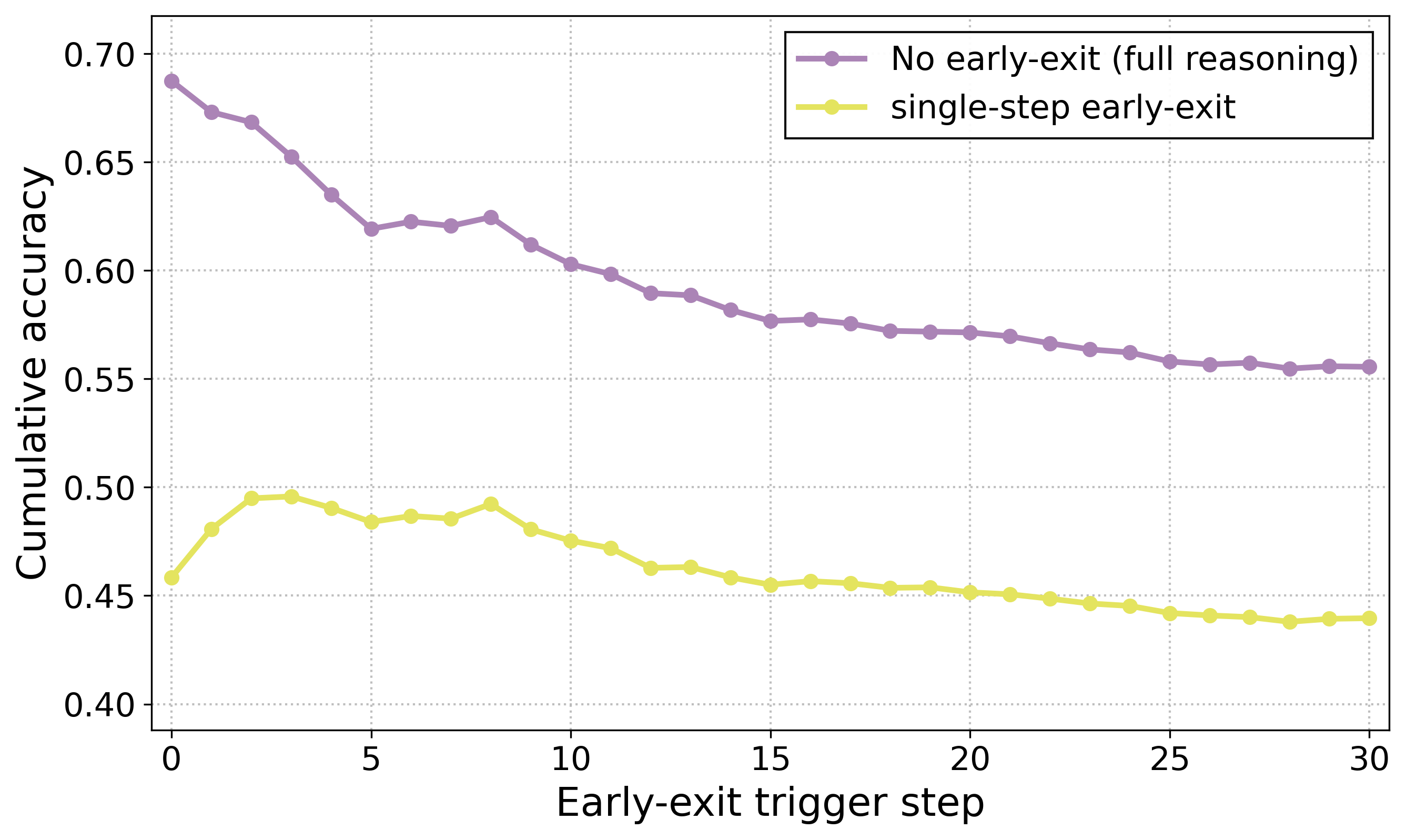}
    \vspace{-1.5em}
    \caption{
    Cumulative accuracy of single-step early-exit based on confidence for DeepSeek-R1-Distill-Llama-8B on hard mathematical reasoning tasks.
    }
    \vspace{-1.2em}
    \label{fig:single_step_early_exit}
\end{figure}

\begin{figure*}[t]
    \centering
    \includegraphics[width=\textwidth]{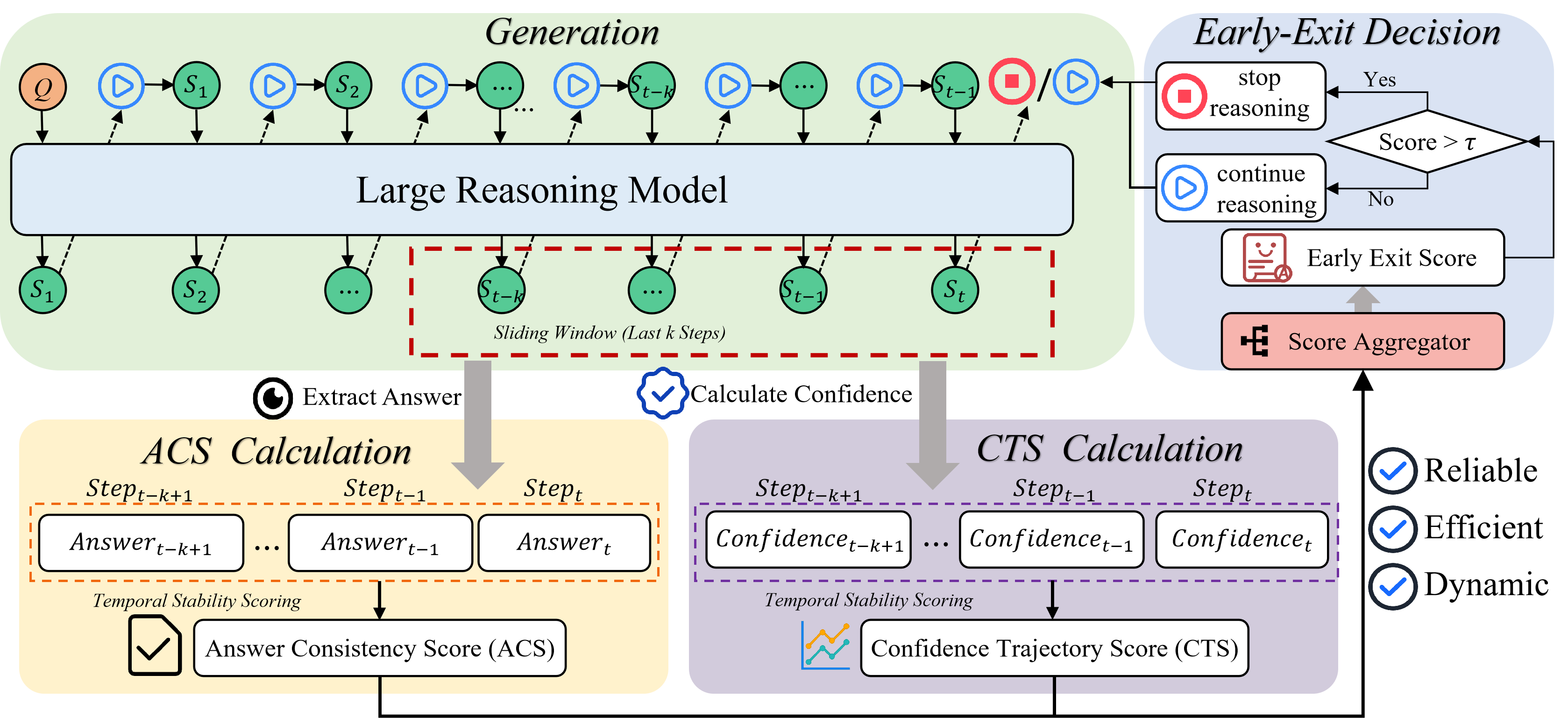}
    \caption{Overview of TRACE.
The model generates reasoning steps autoregressively.
A sliding window over the most recent $k$ steps aggregates answer consistency and confidence trajectory signals, which are jointly used to determine whether the reasoning process has converged and can be safely terminated.}
    \label{fig:our_method}
\end{figure*}

We conduct an empirical evaluation of single-step confidence-based early-exit on hard mathematical reasoning benchmarks, including OlympiadBench, MATH500, AIME24, and AIME25.
Figure~\ref{fig:single_step_early_exit} reports the cumulative accuracy of early-exit decisions as a function of the reasoning step at which exit is triggered, compared to full reasoning.
Across reasoning steps, early-exit decisions based on single-step confidence underperform full reasoning, achieving an overall accuracy of 0.44 compared to 0.55 obtained by full reasoning.

These observations indicate that reliable early-exit decisions require moving beyond instantaneous confidence signals.
In multi-step reasoning, convergence is inherently temporal and is better reflected by the stability of model behavior across steps, motivating the aggregation of evidence over multiple reasoning steps.

\section{Method}

Motivated by the limitations of single-step confidence discussed above, we present \textbf{Temporal Reasoning Aggregation for Convergent Exit (TRACE)}. Instead, TRACE determines when to terminate reasoning by aggregating evidence across a sliding window of recent reasoning steps, capturing whether the model’s predicted answers have stabilized and whether its confidence has evolved consistently over time.
As illustrated in Figure~\ref{fig:our_method}, TRACE operationalizes this principle through two complementary signals: (i) the \textit{Answer Consistency Score (ACS)}, which measures multi-step persistence of induced answers, and (ii) the \textit{Confidence Trajectory Score (CTS)}, which summarizes the temporal evolution of model confidence.
These signals are jointly used by an early-exit decision module to determine when reasoning has genuinely converged and inference can be safely terminated.

\subsection{Multi-Step Evidence Aggregation}
We consider an autoregressive reasoning process in which a large language model generates a sequence of reasoning steps.
Each reasoning step corresponds to a contiguous segment of the generated reasoning that reflects a locally coherent stage of the model’s thought process.
At each step $t$, the model outputs an intermediate reasoning segment together with an associated confidence estimate derived from the model’s generation probabilities.
Our objective is to determine whether reasoning has sufficiently converged to allow safe termination at step $t$, or whether additional reasoning steps should be generated.

To capture temporal dependencies in the reasoning process, TRACE maintains a sliding window over the most recent $k$ reasoning steps.
This window captures a short temporal history of the model's reasoning path, including the predicted answers and confidence values from steps $t-k+1$ to $t$.
Within the sliding window, TRACE aggregates two complementary signals to capture reasoning stability: answer consistency and confidence trajectory.

\paragraph{Answer Consistency Score.}
The ACS measures the stability of predicted answers across recent reasoning steps.
At each step, we induce a candidate final answer based on the reasoning content generated so far.
The induction is implemented via a lightweight auxiliary prompting procedure that reuses the existing reasoning context without interrupting the ongoing reasoning process.
Implementation details are provided in Appendix~\ref{app:answer_induction}.

Given a sliding window of the last $k$ steps, ACS is defined for each candidate answer $a$ as
\begin{equation}
\mathrm{ACS}(a) = \frac{\mathrm{count}(a)}{k},
\label{eq:acs_score}
\end{equation}
where $\mathrm{count}(a)$ denotes the number of times answer $a$ appears within the window.
Intuitively, when reasoning has converged, correct answers tend to reappear persistently across consecutive steps, leading to higher consistency scores.

\paragraph{Confidence Trajectory Score.}
The CTS summarizes the temporal evolution of model confidence associated with a candidate answer across recent reasoning steps.
At each step, we compute a scalar confidence score for the induced candidate answer based on the model’s token-level probability distribution.

Specifically, given the probability distribution over a set of candidate tokens at each position of the induced answer, we quantify uncertainty using a normalized entropy.
For a token with probability distribution $\mathbf{p}$ over its candidate set, the normalized entropy is defined as
\begin{equation}
\tilde{H}(\mathbf{p}) = \frac{-\sum_i p_i \log p_i}{\log |\mathbf{p}|},
\label{eq:norm_entropy}
\end{equation}

where $|\mathbf{p}|$ denotes the number of candidate tokens.
The confidence score for an answer is then computed as
\begin{equation}
c = 1 - \frac{1}{n} \sum_{j=1}^{n} \tilde{H}(\mathbf{p}_j),
\label{eq:confidence_caculate}
\end{equation}
where $n$ is number of tokens in the induced answer and $\mathbf{p}_j$ is the token-level distribution at position $j$.
This formulation assigns higher confidence to answers with more concentrated probability mass and penalizes diffuse or uncertain predictions.

Given a sliding window of last $k$ steps, CTS is defined for each candidate answer $a$ as the average confidence over the steps in which $a$ appears:
\begin{equation}
\mathrm{CTS}(a) = \frac{1}{\mathrm{count}(a)} \sum_{t \in \mathcal{T}(a)} c_t,
\label{eq:cts_score}
\end{equation}
where $\mathcal{T}(a)$ denotes the set of steps within the window at which answer $a$ is predicted, and $c_t$ is the confidence score at step $t$.
By aggregating confidence values over time, CTS distinguishes sustained confidence indicative of reasoning convergence from step-level confidence variations that are insufficient for reliable stopping decisions.

\subsection{Early-Exit Decision}
At each reasoning step, TRACE makes an early-exit decision by jointly considering answer consistency and confidence trajectory signals aggregated over the sliding window.
For each candidate answer $a$ appearing in the window, we compute a unified stability score by combining its Answer Consistency Score and Confidence Trajectory Score:
\begin{equation}
S(a) = \alpha \cdot \mathrm{ACS}(a) + (1 - \alpha) \cdot \mathrm{CTS}(a),
\label{eq:sa_caculate}
\end{equation}
where $\alpha \in [0,1]$ controls the relative contribution of the two signals.

At each reasoning step, we select the candidate answer with the highest stability score,
\begin{equation}
a^\star = \arg\max_a S(a).
\label{eq:argmax}
\end{equation}
If $S(a^\star)$ exceeds a predefined threshold $\tau$, the reasoning process is considered to have converged, and inference is terminated with $a^\star$ as the final answer.
Otherwise, the model continues generating subsequent reasoning steps.

This early-exit decision is applied dynamically at inference time and requires no additional training or modification to the underlying model, making TRACE readily applicable to off-the-shelf LLMs.

\section{Experiments}
% Required packages:
% \usepackage{booktabs}
% \usepackage{multirow}

% Preamble needs:
% \usepackage{booktabs}

\begin{table*}[t]
\centering
\footnotesize
\setlength{\tabcolsep}{1.9pt}
\renewcommand{\arraystretch}{1.06}

\begin{tabular}{l
ccc ccc ccc ccc ccc | cc}
\toprule
\textbf{Method}
& \multicolumn{3}{c}{\textbf{OlympiadBench}}
& \multicolumn{3}{c}{\textbf{MATH500}}
& \multicolumn{3}{c}{\textbf{AIME24}}
& \multicolumn{3}{c}{\textbf{AMC23}}
& \multicolumn{3}{c}{\textbf{AIME25}}
& \multicolumn{2}{c}{\textbf{AVG}} \\
\cmidrule(lr){2-4}\cmidrule(lr){5-7}\cmidrule(lr){8-10}
\cmidrule(lr){11-13}\cmidrule(lr){14-16}\cmidrule(lr){17-18}
& Acc$\!\uparrow$ & Tok$\downarrow$ & CR$\downarrow$
& Acc$\!\uparrow$ & Tok$\downarrow$ & CR$\downarrow$
& Acc$\!\uparrow$ & Tok$\downarrow$ & CR$\downarrow$
& Acc$\!\uparrow$ & Tok$\downarrow$ & CR$\downarrow$
& Acc$\!\uparrow$ & Tok$\downarrow$ & CR$\downarrow$
& Acc$\!\uparrow$ & CR$\downarrow$ \\

\midrule

\multicolumn{18}{l}{\textbf{Qwen3-8B}}\\
Vanilla
& 81.0 & 10518 & 100\%
& 95.2 & 5138  & 100\%
& 70.3 & 14477 & 100\%
& 96.6 & 7885  & 100\%
& 59.6 & 16604 & 100\%
& 80.5 & 100\% \\

TALE
& 73.1 & 8417  & 80.0\%
& 93.1 & 3544  & 69.0\%
& 68.1 & 10166 & 70.2\%
& 94.1 & 5890  & 74.7\%
& 55.7 & 12822 & 77.2\%
& 76.8 & 74.2\% \\

Dynasor
& 77.3 & 9148  & 87.0\%
& 93.7 & 3873  & 75.4\%
& 66.7 & 12638 & 87.3\%
& 95.9 & 6552  & 83.1\%
& 53.1 & 13447 & 81.0\%
& \underline{77.3} & 82.8\% \\

NoThink
& 57.2 & 1710  & 16.3\%
& 84.2 & 923   & 18.0\%
& 39.7 & 3812  & 26.3\%
& 74.8 & 1575  & 20.0\%
& 21.0 & 2844  & 17.1\%
& 55.3 & \textbf{19.5\%} \\

DEER
& 74.2 & 8074  & 76.8\%
& 92.9 & 3468  & 67.5\%
& 66.3 & 9645  & 66.6\%
& 94.7 & 6455  & 81.9\%
& 53.3 & 12453 & 75.0\%
& 76.3 & 73.6\% \\

\textbf{TRACE}
& 80.0 & 8088  & 76.9\%
& 94.1 & 3231  & 62.9\%
& 68.8 & 10532 & 72.8\%
& 96.0 & 5405  & 68.6\%
& 57.5 & 11512 & 69.3\%
& \textbf{79.3} & \underline{70.1\%} \\
\midrule

\multicolumn{18}{l}{\textbf{Qwen3-4B}}\\
Vanilla
& 77.9 & 10294 & 100\%
& 93.6 & 5229  & 100\%
& 64.7 & 14318 & 100\%
& 93.7 & 8089  & 100\%
& 54.7 & 16580 & 100\%
& 76.9 & 100\% \\

TALE
& 69.5 & 8521  & 82.8\%
& 89.7 & 2938  & 56.2\%
& 57.3 & 9373  & 65.5\%
& 93.1 & 6772  & 83.7\%
& 54.3 & 10411 & 62.8\%
& 72.8 & \underline{70.2\%} \\

Dynasor
& 73.8 & 9133  & 88.7\%
& 91.4 & 4339  & 83.0\%
& 62.3 & 12755 & 89.1\%
& 90.1 & 6812  & 84.2\%
& 52.1 & 13816 & 83.3\%
& \underline{73.9} & 85.7\% \\

NoThink
& 55.0 & 1804  & 17.5\%
& 80.4 & 908   & 17.4\%
& 23.3 & 5459  & 38.1\%
& 72.5 & 1814  & 22.4\%
& 16.7 & 3417  & 20.6\%
& 49.6 & \textbf{23.2\%} \\

DEER
& 74.7 & 9696  & 94.2\%
& 92.8 & 3944  & 75.4\%
& 52.7 & 8720  & 60.9\%
& 93.5 & 6588  & 81.5\%
& 53.1 & 14160 & 85.4\%
& 73.4 & 79.5\% \\

\textbf{TRACE}
& 77.6 & 9489 & 92.2\%
& 93.2 & 3788 & 72.4\%
& 61.5 & 9204 & 64.3\%
& 92.7 & 5652 & 69.9\%
& 53.3 & 11928 & 71.9\%
& \textbf{75.7} & 74.1\% \\
\midrule

\multicolumn{18}{l}{\textbf{Gemini2.5-Flash}}\\
Vanilla
& 77.6 & 6684  & 100\%
& 92.8 & 1449  & 100\%
& 78.2 & 8476  & 100\%
& 98.8 & 2290  & 100\%
& 66.3 & 11993 & 100\%
& 82.7 & 100\% \\

TALE
& 68.4 & 4950  & 74.0\%
& 91.9 & 1411  & 97.4\%
& 69.1 & 7361  & 86.8\%
& 92.3 & 1856  & 81.0\%
& 54.3 & 9162  & 76.4\%
& 75.2 & 83.1\% \\

Dynasor
& 72.7 & 5919  & 88.5\%
& 92.0 & 1311  & 90.5\%
& 74.7 & 6976  & 82.3\%
& 95.3 & 2001  & 87.4\%
& 62.6 & 10178 & 84.9\%
& 79.5 & 86.7\% \\

NoThink
& 74.1 & 6296  & 94.2\%
& 90.6 & 1400  & 96.6\%
& 76.3 & 8214  & 96.9\%
& 96.5 & 2311  & 101\%
& 64.7 & 12221 & 102\%
& \underline{80.4} & 98.1\% \\

DEER
& 70.3 & 5805  & 86.8\%
& 91.2 & 1224  & 84.5\%
& 66.7 & 6326  & 74.6\%
& 93.1 & 1874  & 81.8\%
& 56.7 & 9271  & 77.3\%
& 75.6 & \underline{81.0\%} \\

\textbf{TRACE}
& 75.3 & 4291 & 64.2\%
& 92.2 & 1166 & 80.5\%
& 74.9 & 6897 & 81.4\%
& 98.5 & 1994 & 87.4\%
& 62.6 & 9219 & 76.9\%
& \textbf{80.7} & \textbf{78.0\%} \\
\midrule

\multicolumn{18}{l}{\textbf{R1-Distilled-Llama-8B}}\\
Vanilla
& 58.2 & 6286  & 100\%
& 87.2 & 2934  & 100\%
& 43.3 & 12648 & 100\%
& 91.7 & 5717  & 100\%
& 32.9 & 10828 & 100\%
& 62.7 & 100\% \\

TALE
& 56.2 & 5713  & 90.9\%
& 85.3 & 2455  & 83.7\%
& 32.1 & 8724  & 69.0\%
& 82.5 & 3169  & 55.4\%
& 26.0 & 9139  & 84.4\%
& 56.4 & \textbf{76.7\%} \\

Dynasor
& 56.9 & 5814  & 92.5\%
& 86.4 & 2693  & 91.8\%
& 42.2 & 10912 & 86.3\%
& 89.3 & 4811  & 84.2\%
& 30.0 & 9634  & 89.0\%
& \underline{61.0} & 88.7\% \\

NoThink
& 56.0 & 6150  & 97.8\%
& 85.2 & 3024  & 103\%
& 41.3 & 12803 & 101\%
& 88.1 & 5288  & 92.5\%
& 31.0 & 10386 & 95.9\%
& 60.3 & 98.1\% \\

DEER
& 55.4 & 5443  & 86.6\%
& 86.4 & 2718  & 92.6\%
& 34.5 & 9070  & 71.7\%
& 84.1 & 4551  & 79.6\%
& 29.3 & 9755  & 90.1\%
& 57.9 & 84.1\% \\

\textbf{TRACE}
& 57.5 & 4784 & 76.1\%
& 85.4 & 2436 & 83.0\%
& 41.9 & 9858 & 77.9\%
& 88.5 & 4576 & 80.1\%
& 32.7 & 8302 & 76.7\%
& \textbf{61.2} & \underline{78.8\%} \\

\bottomrule
\end{tabular}
\caption{Main results on mathematical reasoning benchmarks.
Acc denotes accuracy, Tok denotes token count, and CR denotes compression rate relative to Vanilla. $\uparrow$ ($\downarrow$) indicates higher (lower) values are better. \textbf{Bold} and \underline{underlined} numbers indicate the best and second-best results among all methods, respectively.
Vanilla full reasoning is used as the full-compute baseline (CR$=100\%$ by definition) and is not included in the ranking, since it does not perform early exit and is not directly comparable in efficiency.
}

\vspace{-1.2em}
\label{tab:main_results}
\end{table*}

\begin{table}[t]
\centering
\footnotesize
\setlength{\tabcolsep}{2.8pt}
\renewcommand{\arraystretch}{1.12}

\resizebox{\linewidth}{!}{%
\begin{tabular}{@{}l c c c @{\quad} l c c c@{}}
\toprule
Method & Acc$\uparrow$ & Tok$\downarrow$ & CR$\downarrow$ & Method & Acc$\uparrow$ & Tok$\downarrow$ & CR$\downarrow$ \\
\midrule

\multicolumn{8}{l}{\textbf{Qwen3-8B}}\\
Vanilla & 59.1 & 9398 & 100\%
& TALE & 50.0 & \underline{4245} & \underline{45.2\%} \\
Dynasor & \underline{54.6} & 5732 & 61.0\%
& NoThink & 51.0 & \textbf{1538} & \textbf{16.4\%} \\
DEER & 53.2 & 7279 & 77.5\%
& TRACE & \textbf{56.6} & 5629 & 59.9\% \\
\midrule

\multicolumn{8}{l}{\textbf{Qwen3-4B}}\\
Vanilla & 55.1 & 8611 & 100\%
& TALE & 47.9 & 5807 & 67.4\% \\
Dynasor & 52.1 & \underline{4814} & \underline{55.9\%}
& NoThink & 49.0 & \textbf{1658} & \textbf{19.3\%} \\
DEER & \underline{54.5} & 7042 & 81.8\%
& TRACE & \textbf{55.0} & 5713 & 67.3\% \\
\midrule

\multicolumn{8}{l}{\textbf{Gemini2.5-Flash}}\\
Vanilla & 69.5 & 4565 & 100\%
& TALE & 68.2 & 3597 & 78.8\% \\
Dynasor & \textbf{69.5} & 4091 & 89.6\%
& NoThink & 67.3 & 4365 & 95.6\% \\
DEER & 69.1 & \underline{3397} & \underline{74.4\%}
& TRACE &  \underline{69.4} & \textbf{3262} &  \textbf{71.4\%} \\
\midrule

\multicolumn{8}{l}{\textbf{R1-Distilled-Llama-8B}}\\
Vanilla & 51.0 & 6184 & 100\%
& TALE & 37.4 & \underline{4531} & \underline{73.3\%} \\
Dynasor & 43.6 &  \textbf{4019} &  \textbf{65.0\%}
& NoThink & \underline{50.5} & 5849 & 94.6\% \\
DEER & 43.0 & 4731 & 76.5\%
& TRACE &  \textbf{51.5} & 4644 & 75.1\% \\
\bottomrule
\end{tabular}}

\caption{Results on the GPQA-D science benchmark.}
\vspace{-1.2em}
\label{tab:gpqa}
\end{table}

\subsection{Experimental Setup}

\paragraph{Datasets.}
We evaluate our method on a collection of challenging mathematical and scientific reasoning benchmarks,
including OlympiadBench \citep{he2024olympiadbench}, MATH500 \citep{lightman2023let}, AIME24, AIME25 \citep{aime_problems_solutions}, AMC23 \citep{aimo_validation_amc}, and the GPQA-D science benchmark \citep{rein2024gpqa}.
These benchmarks require multi-step reasoning and often involve long chain-of-thought generation,
making them well-suited for evaluating early-exit strategies under test-time scaling.
All experiments are conducted in a zero-shot setting using the official evaluation protocols of each benchmark.

\paragraph{Models.}
We conduct experiments on a diverse set of LLMs spanning multiple scales and architectures, including Qwen3-8B, Qwen3-4B \citep{yang2025qwen3technicalreport}, Gemini-2.5-Flash \citep{comanici2025gemini}, and DeepSeek-R1-Distilled-Llama-8B \citep{guo2025deepseek}, covering both open- and closed-source systems.
For each model, we adopt the recommended reasoning prompts and decoding configurations.

\paragraph{Baselines.}
We compare TRACE with representative baselines that mitigate overthinking either by constraining the length of reasoning or by terminating reasoning early.
\textbf{Vanilla full reasoning} generates the full chain-of-thought trace until completion and serves as the reference for accuracy and token cost.
\textbf{NoThinking} is a model-specific prompting intervention that suppresses the reasoning phase.
\textbf{TALE} explicitly constrains the length of the reasoning trace by a problem-specific token budget. 
For adaptive early exit, \textbf{Dynasor} leverages a single probe-based indicator derived from answer stabilization to allocate token budgets. 
\textbf{DEER} performs single-step early exit by applying a single-step confidence criterion to decide termination. 
All methods are evaluated using accuracy (Acc, $\uparrow$), average generated tokens (Tok, $\downarrow$), and compression rate (CR, $\downarrow$), defined as the relative token reduction compared to vanilla full reasoning.

\paragraph{Implementation Details.}
Our method operates entirely at inference time and requires no additional training.
Reasoning steps are segmented based on discourse-level transition markers that indicate shifts in the model’s line of thought.
We maintain a sliding window of the most recent $k$ reasoning steps to aggregate answer consistency and confidence trajectory signals. Detailed implementation settings and hyperparameter choices are provided in Appendix~\ref{app:Hyper-parameters Setting}.

\subsection{Main Results}
Table~\ref{tab:main_results} reports the main results of dynamic early-exit methods on multiple hard mathematical reasoning benchmarks across different models.
Overall, our method consistently achieves higher accuracy than existing early-exit baselines while substantially reducing inference cost in most cases.
Across all evaluated models, the average accuracy drop relative to full reasoning is within 2\%, indicating that our method preserves most of the original reasoning correctness despite early termination.
At the same time, our approach reduces the number of generated tokens by approximately 25\% on average across models and datasets, demonstrating effective mitigation of overthinking with minimal accuracy loss.

Compared to prior confidence-based early-exit methods such as TALE and DEER,
TRACE yields a more favorable accuracy--efficiency trade-off.
While TALE aggressively reduces token usage at the cost of significant accuracy degradation,
our approach maintains markedly higher accuracy while achieving comparable or lower token consumption.
In most settings, our method outperforms DEER in accuracy under similar or lower compute ratios,
demonstrating more reliable early-exit decisions.

Beyond mathematical reasoning benchmarks, we evaluate TRACE on the GPQA-D science dataset to assess its generalization beyond math.
As shown in Table~\ref{tab:gpqa}, TRACE reduces inference tokens by approximately 20--30\% across all evaluated models with less accuracy drop compared to full reasoning.
This suggests that TRACE does not rely on domain-specific heuristics and can generalize effectively beyond mathematical settings.

Across both mathematical and scientific reasoning tasks, the observed trends hold consistently across diverse model families,
including open-source models of different scales as well as the closed-source Gemini model,
highlighting the robustness and general applicability of our approach.

\subsection{Comparison with Single-Step Confidence Early-Exit}
\begin{figure}[t]
    \centering
    \includegraphics[width=\linewidth]{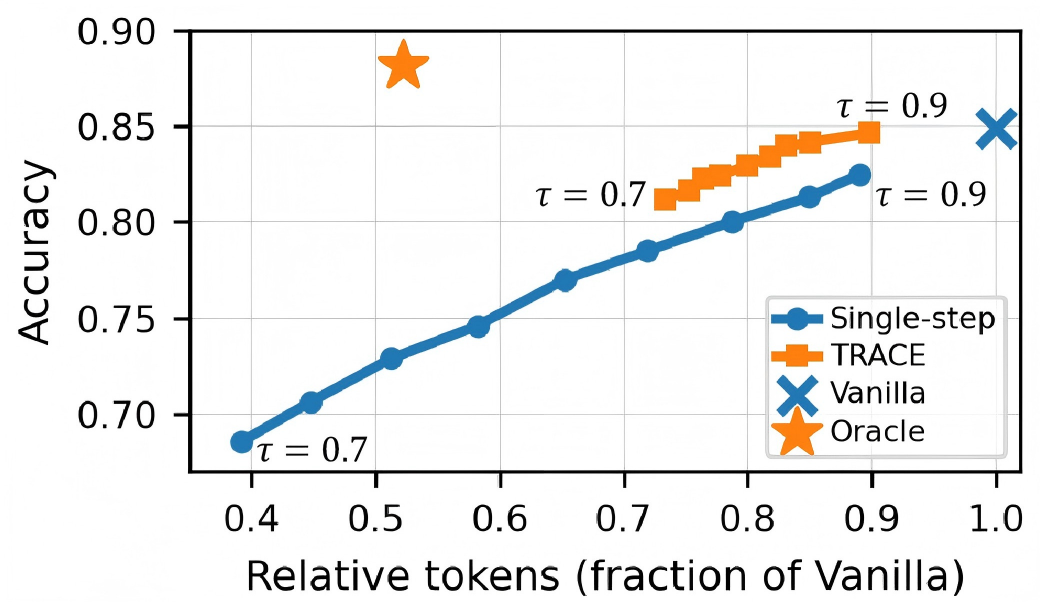}
    \vspace{-2em}
    \caption{Accuracy--token trade-off under identical early-exit thresholds. Each point on the curves corresponds to a different early exit threshold $\tau$. Oracle denotes an idealized upper bound that stops at the first step where the induced answer is correct.}
    \vspace{-1.2em}
    \label{fig:early-stop-tradeoff}
\end{figure}

Figure~\ref{fig:early-stop-tradeoff} presents a comparison between TRACE and single-step confidence-based early exit under identical stopping thresholds $\tau$ on hard mathematical reasoning benchmarks. Results are aggregated across all math datasets using Qwen3-4B.

As shown in the figure, single-step confidence exhibits high sensitivity to threshold selection, with accuracy varying as the threshold changes.
In contrast, TRACE demonstrates robustness to threshold perturbations as the achieved accuracy remains nearly constant across a wide range of thresholds.
This suggests that aggregating multi-step evidence yields a more stable early-exit signal than relying on instantaneous confidence alone.

Moreover, TRACE consistently achieves a better accuracy--efficiency trade-off than the single-step baseline.
From the x-axis perspective (token usage), TRACE requires fewer generated tokens to reach a comparable accuracy level, indicating that it mitigates delayed termination and reduces redundant computation.
From the y-axis perspective (accuracy), TRACE attains higher accuracy under the same token budget, suggesting that it is less prone to premature termination caused by confidence fluctuations.

\section{Analysis}

\begin{figure*}[t]
  \centering
  \begin{minipage}[t]{0.62\textwidth}
    \centering
    \includegraphics[width=\linewidth]{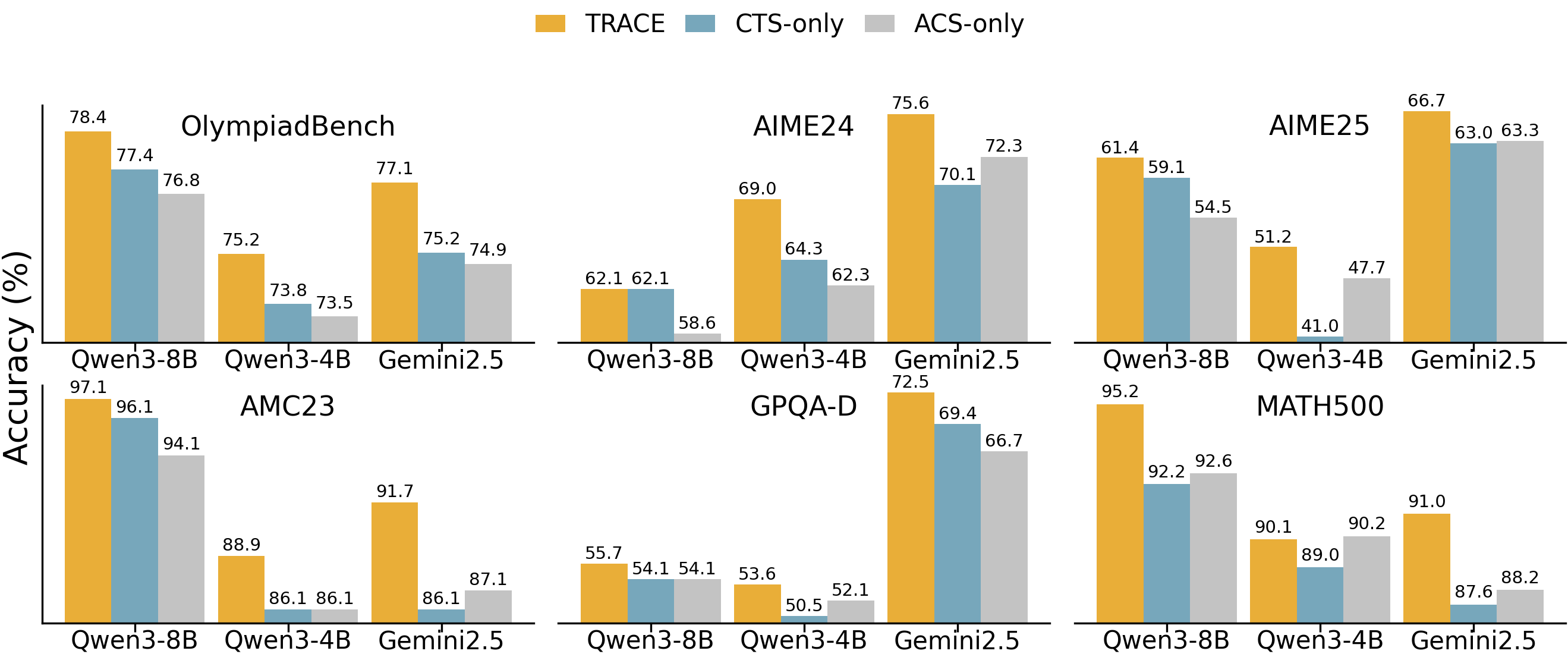}
    \captionof{figure}{%
    Accuracy (\%) of TRACE early-exit across multiple math benchmarks and models: ablation of ACS and CTS (ACS+CTS (TRACE, our full method) vs. ACS-only vs. CTS-only) under matched token budgets on triggered examples.
    }
    \vspace{-1.2em}
    \label{fig:ablation_acs_cts}
  \end{minipage}\hfill
  \begin{minipage}[t]{0.36\textwidth}
    \centering
    \includegraphics[width=\linewidth]{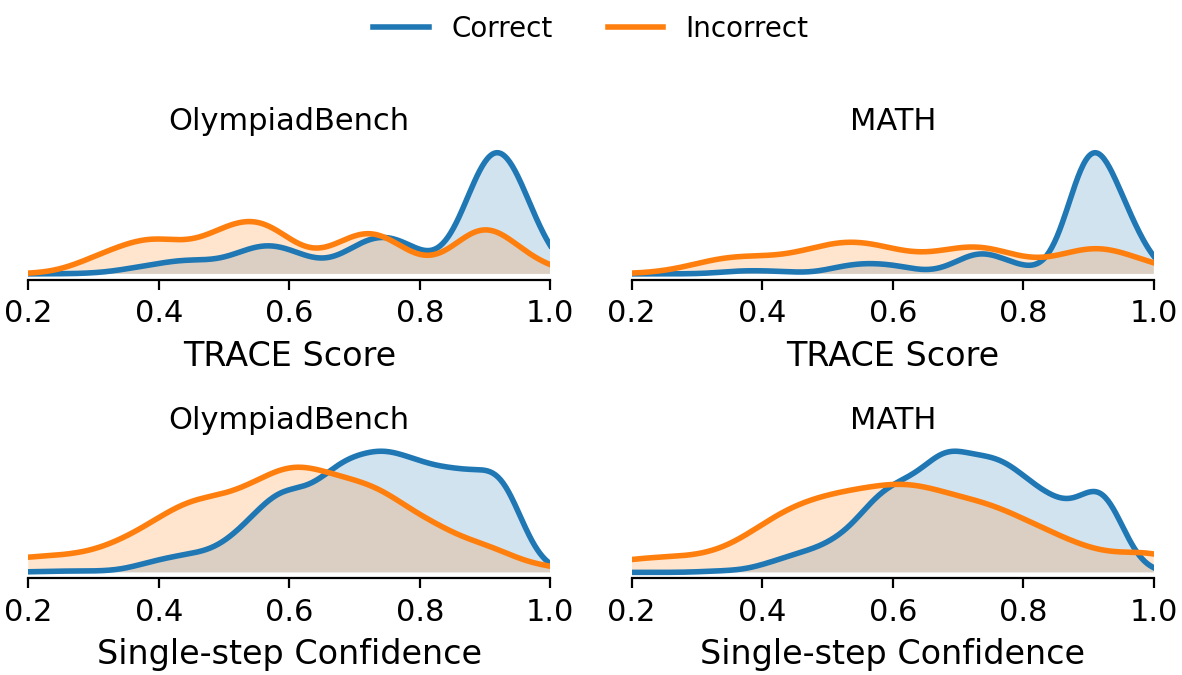}
    \captionof{figure}{%
    Kernel density estimation of TRACE stability scores and single-step confidence for Qwen3-8B on OlympiadBench and MATH.
    }
    \vspace{-1.2em}
    \label{fig:conf_kde_qwen8b}
  \end{minipage}
\end{figure*}

\subsection{Ablation Study}
To isolate the contribution of each component in TRACE, we ablate the early-exit criterion by comparing three variants: (i) \textbf{ACS+CTS} (our full method),
(ii) \textbf{CTS-only}, which triggers early exit using aggregated confidence signals only, and
(iii) \textbf{ACS-only}, which relies solely on answer consistency.
Figure~\ref{fig:ablation_acs_cts} reports accuracy for three models across six benchmarks, evaluated on examples where early exit is triggered. All settings use matched token budgets to ensure fair comparison.

Across all evaluated benchmarks, TRACE (ACS+CTS) is consistently the best or tied-best variant. However, both single-component variants degrade on harder datasets (e.g., AIME), whereas combining ACS and CTS yields robust improvements. 
This demonstrates their complementarity: ACS prevents premature exit by enforcing cross-step agreement but can be fooled by consistently wrong answers, whereas CTS reflects growing certainty but may trigger too early when confidence increases before answers stabilize; combining both yields a more reliable convergence signal.

\subsection{TRACE Score Distributions}

We compare TRACE stability scores with single-step confidence in separating correct from incorrect steps. Figure~\ref{fig:conf_kde_qwen8b} plots their step-level distributions for Qwen3-8B on OlympiadBench and MATH. A correct step denotes a reasoning step whose induced answer is correct, while an incorrect step denotes a step whose induced answer is incorrect.

Single-step confidence is informative but noisy. Although correct steps shift toward higher values, the correct and incorrect distributions still overlap with incorrect steps showing noticeable mass at high confidence. This proves single-step confidence is too noisy to admit a threshold that reliably balances accuracy and efficiency.

In contrast, TRACE yields a reliable high-score signal by requiring answer consistency across steps, making it less sensitive to transient confidence spikes. Correct steps concentrate near the upper end of the TRACE range (often peaking around 0.8--1.0), while incorrect steps show much less mass there and remain flatter and more diffuse. This produces a cleaner separation in the high-score regime where early-exit decisions are made, enabling high-threshold exit with lower risk of premature exits.

\begin{figure}[t]
  \centering
  \begin{subfigure}[b]{0.48\linewidth}
    \centering
    \includegraphics[width=\linewidth]{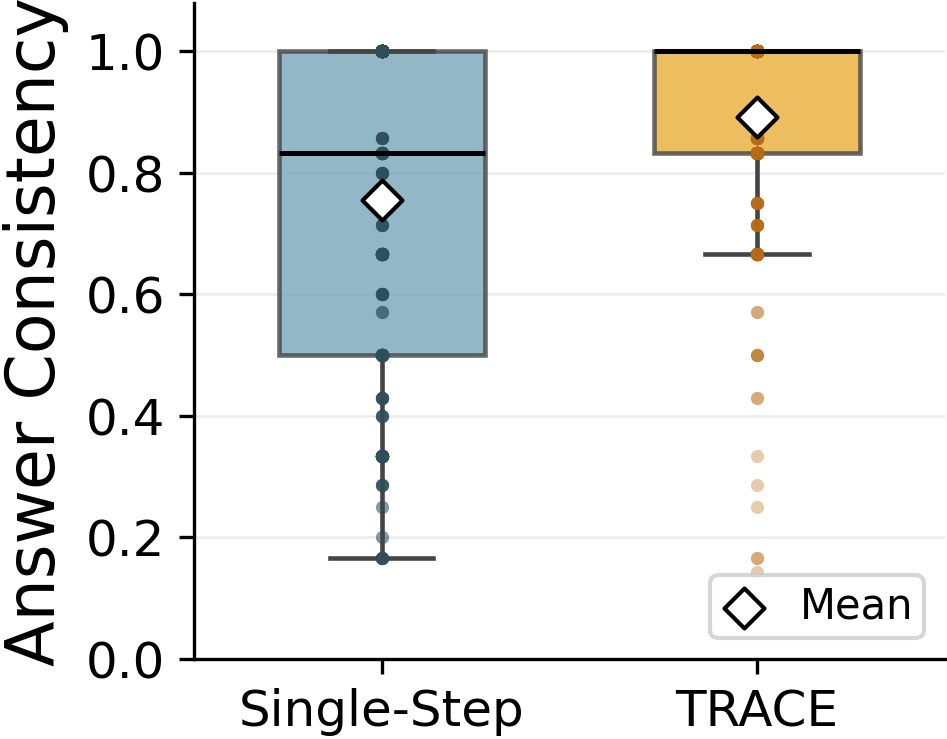}
    \caption{Qwen3-8B}
    \label{fig:consistency-qwen3-8b}
  \end{subfigure}\hfill
  \begin{subfigure}[b]{0.48\linewidth}
    \centering
    \includegraphics[width=\linewidth]{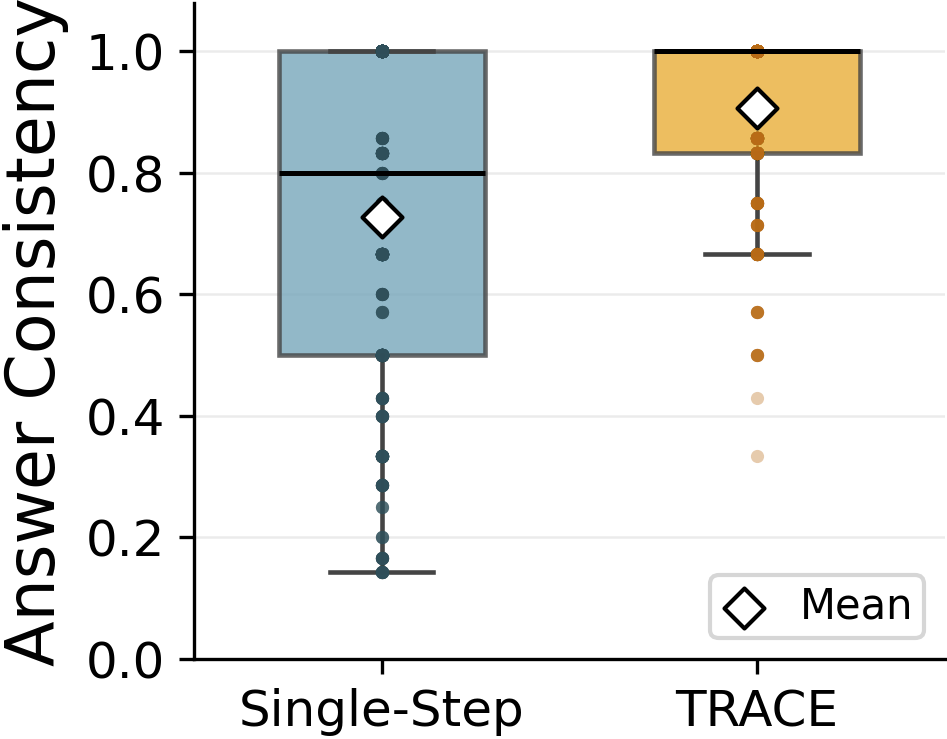}
    \caption{Qwen3-4B}
    \label{fig:consistency-qwen3-4b}
  \end{subfigure}

  \vspace{1mm}

  \begin{subfigure}[b]{0.48\linewidth}
    \centering
    \includegraphics[width=\linewidth]{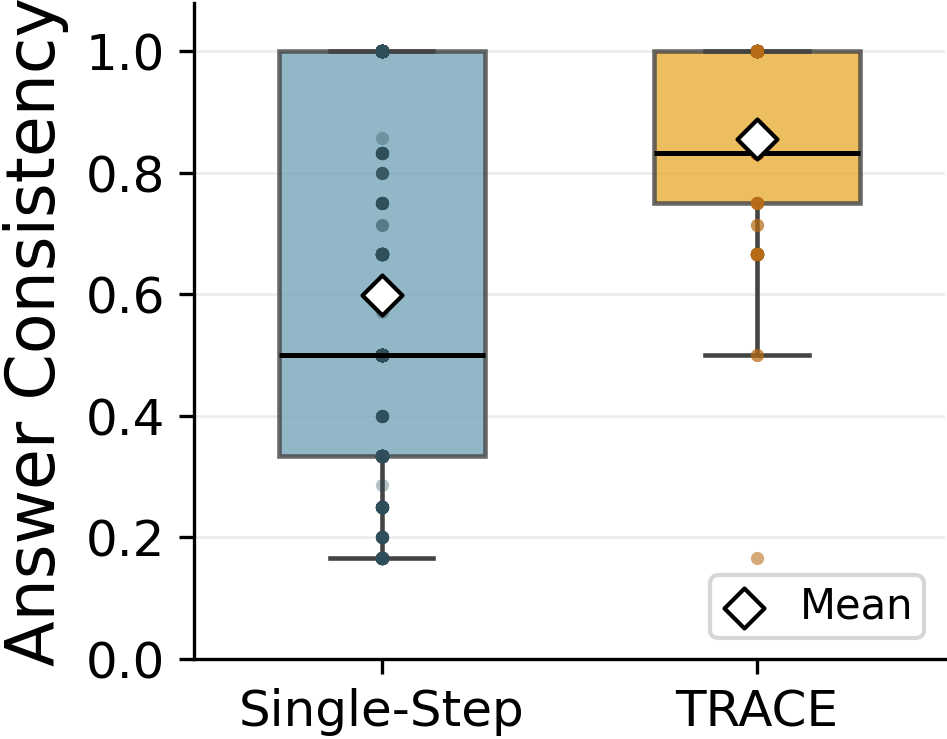}
    \caption{Gemini-2.5}
    \label{fig:consistency-gemini}
  \end{subfigure}\hfill
  \begin{subfigure}[b]{0.48\linewidth}
    \centering
    \includegraphics[width=\linewidth]{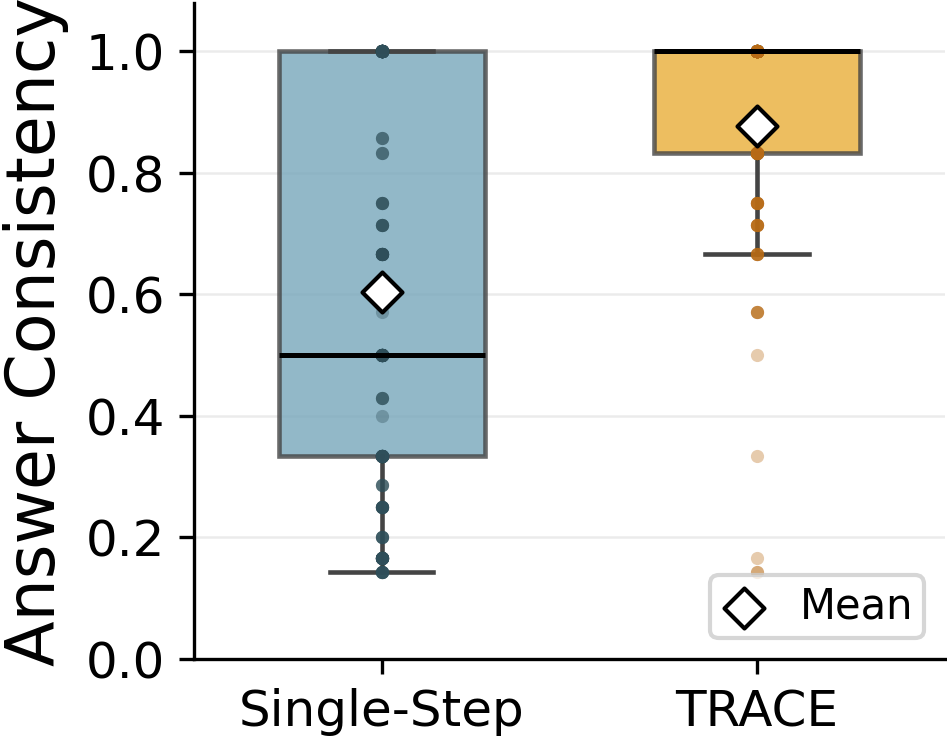}
    \caption{R1-Distill-llama}
    \label{fig:consistency-llama}
  \end{subfigure}

  \caption{Answer consistency comparison (Single-Step vs TRACE) across different backbone models.}
  \vspace{-1.2em}
  \label{fig:consistency-2x2}
\end{figure}

\subsection{Answer Convergence at Early Exit}
To understand where different early-exit strategies terminate, we measure answer consistency, a proxy for whether the reasoning trajectory has converged at the exit point. Answer consistency is defined as how often the final answer appears within the $k$-step window that triggers early exit. Figure~\ref{fig:consistency-2x2} reports the distributions for a single-step confidence baseline and TRACE across backbone models.

Single-step early-exit often stops before the trajectory has fully stabilized: its consistency values are widely spread with a pronounced low-consistency tail, which indicates that exits can occur while recent answers still fluctuate. This indicates that single-step exits can occur before the reasoning trajectory has converged.

In contrast, TRACE exits mostly at highly stabilized states. Across models, TRACE produces higher and tighter consistency distributions, often clustered near $1.0$, implying the same answer is repeatedly produced over consecutive steps before termination. This provides direct evidence that TRACE favors stable reasoning trajectories and is less sensitive to isolated confidence spikes, helping explain its improved early-exit reliability.

\begin{figure}
    \centering
    \includegraphics[width=1\linewidth]{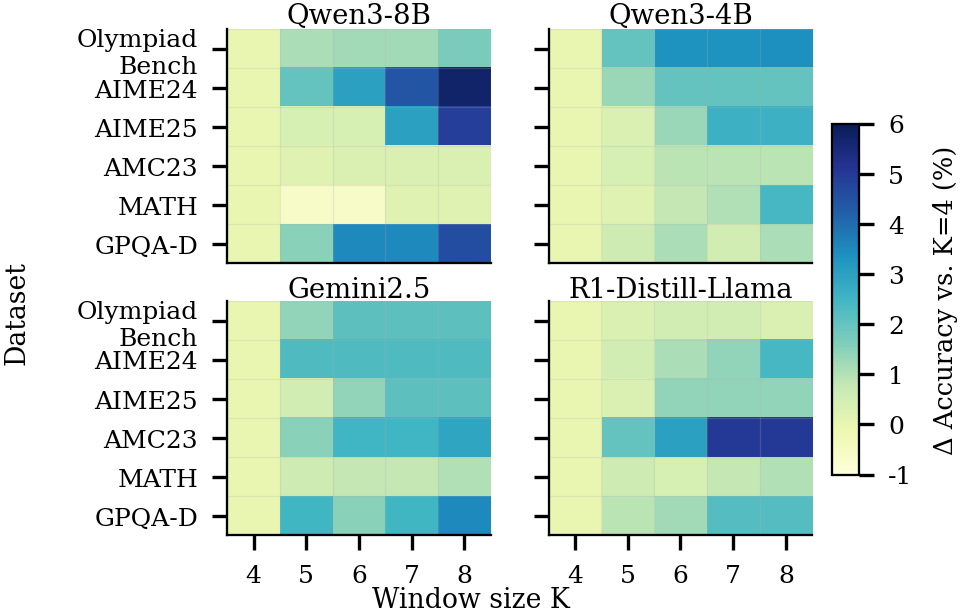}
    \vspace{-1.5em}
    \caption{Sensitivity of TRACE to the window size $k$ across models and benchmarks.}
    \vspace{-0.8em}
    \label{fig:k_sensitivity}
\end{figure}

\subsection{Window Size Robustness}
TRACE aggregates evidence over a window of size $k$, so we evaluate its sensitivity to this hyperparameter. Figure~\ref{fig:k_sensitivity} shows the change in early-exit accuracy when varying $k$ from 4 to 8 across six benchmarks and four backbone models (We report $\Delta$ accuracy relative to $k{=}4$:
$\Delta \mathrm{Acc}_k = \mathrm{Acc}_k - \mathrm{Acc}_4$).
Across all models, accuracy varies smoothly with $k$ and exhibits consistent, mostly monotonic improvements, with little to no degradation as $k$ increases. The gains are more pronounced on harder benchmarks such as AIME and GPQA-D, suggesting that a longer window better suppresses step-level noise and strengthens convergence verification. Importantly, TRACE exhibits limited sensitivity to $k$: even small window sizes perform comparably to larger ones, indicating that TRACE is robust to the window-size choice in practice.

\begin{figure}[t]
    \centering
    \begin{subfigure}[t]{0.49\columnwidth}
        \centering
        \includegraphics[width=\linewidth]{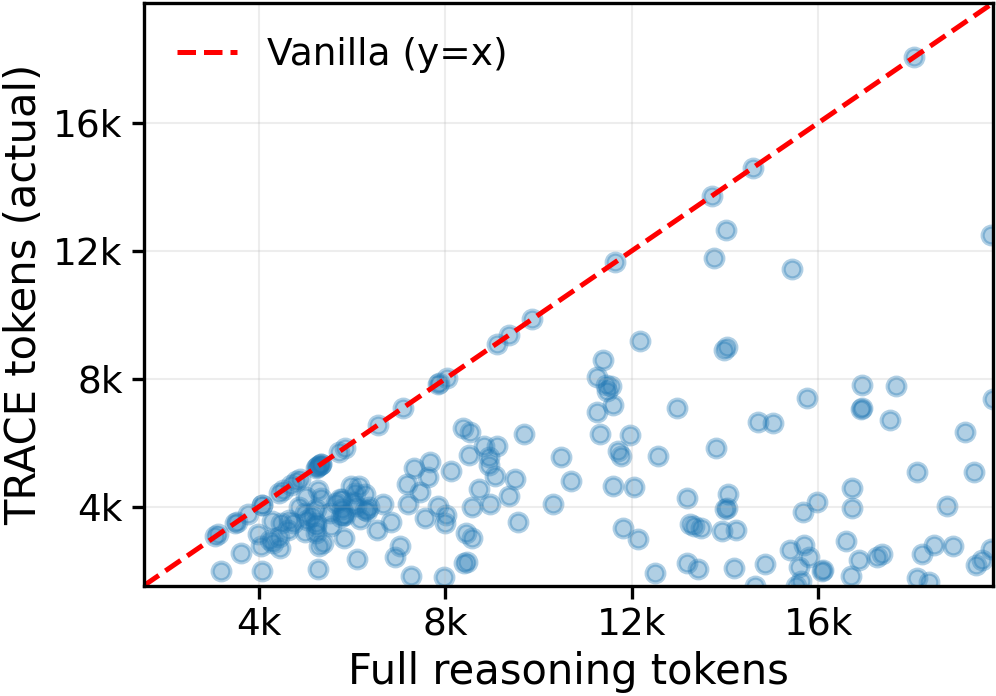}
        \caption{Qwen3-8B}
        \label{fig:cost-scatter-qwen3-8b}
    \end{subfigure}
    \hfill
    \begin{subfigure}[t]{0.49\columnwidth}
        \centering
        \includegraphics[width=\linewidth]{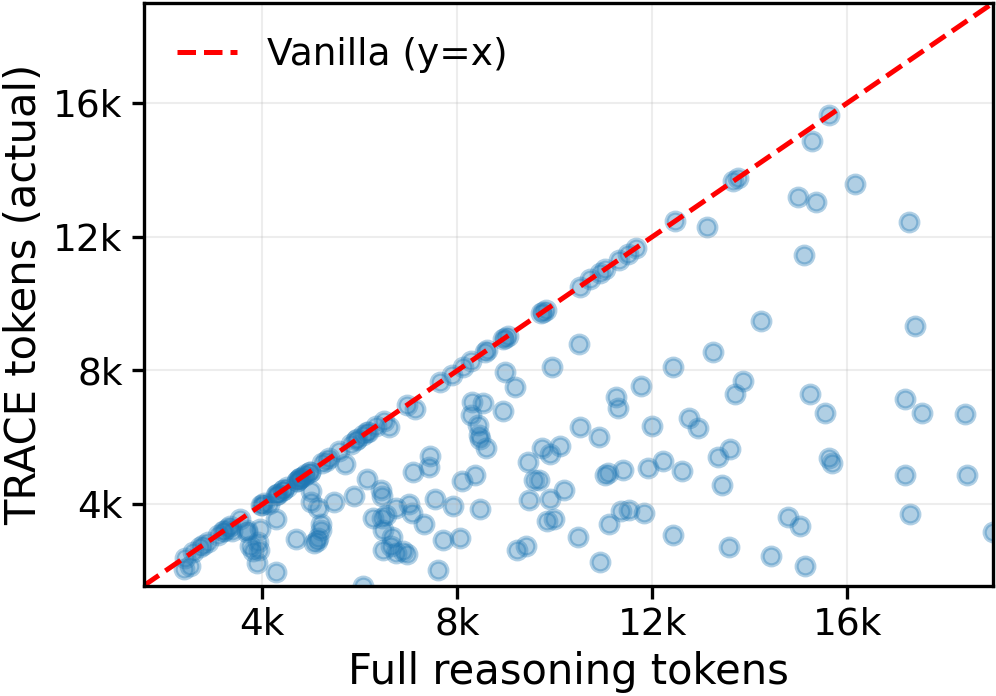}
        \caption{R1-Distill-Llama}
        \label{fig:cost-scatter-r1-llama}
    \end{subfigure}
    
    \vspace{-0.8em}
    \caption{Per-example decoding cost under early stopping on Olympiadbench}
    \vspace{-1.2em}
    \label{fig:cost-scatter-olympiadbench}
\end{figure}

\subsection{Inference Speedup}
Figure~\ref{fig:cost-scatter-olympiadbench} plots the per-example token consumption of TRACE against the corresponding vanilla full reasoning token budget, with the dashed line y=x indicating no savings. Across both Qwen3-8B and R1-Distill-Llama, most points lie well below $y=x$, demonstrating consistent reductions in generated tokens. The gap grows with larger budgets, indicating that early exit yields greater savings on longer reasoning traces where inference cost is dominated by long-tail examples. Overall, TRACE reduce decoding cost and inference time in practice especially for high-budget instances. 

\subsection{Sensitivity to Hyperparameters}
We evaluate the sensitivity of TRACE to the stopping threshold $\tau$ and the mixing weight $\alpha$. 
As shown in Table~\ref{tab:tau_sensitivity}, increasing $\tau$ yields a more conservative stopping policy, resulting in higher accuracy with increased token usage. The trade-off changes consistently across models, and the default choice $\tau=0.8$ provides a good balance between effectiveness and efficiency.
Table~\ref{tab:alpha_sensitivity} shows that larger $\alpha$ (more weight on answer consistency) leads to higher accuracy with higher token cost, while smaller $\alpha$ reduces token usage with modest accuracy degradation. The default setting $\alpha=0.7$ achieves a favorable trade-off.
Overall, TRACE exhibits stable behavior across a range of $\tau$ and $\alpha$, indicating that it does not require delicate hyperparameter tuning.

\begin{table}[t]
\centering
\small
\setlength{\tabcolsep}{5pt}
\begin{tabular}{lccc}
\toprule
Model & $\alpha=0.3$ & $\alpha=0.5$ & $\alpha=0.7$ \\
\midrule
Qwen3-8B & 84.0 / 8.0k & 82.6 / 7.0k & 81.5 / 6.7k \\
Qwen3-4B & 83.1 / 7.5k & 81.8 / 6.5k & 80.7 / 6.2k \\
R1-Distill-Llama & 54.2 / 8.4k & 53.4 / 7.8k & 52.8 / 7.4k \\
Gemini2.5-Flash & 80.7 / 3.5k & 80.5 / 3.4k & 80.2 / 3.3k \\
\bottomrule
\end{tabular}
\caption{Sensitivity of TRACE to the mixing weight $\alpha$ on overall benchmarks, with fixed $k=5$ and $\tau=0.8$. Each entry reports accuracy / average tokens.}
\label{tab:alpha_sensitivity}
\end{table}

\begin{table}[t]
\centering
\small
\setlength{\tabcolsep}{5pt}
\begin{tabular}{lccc}
\toprule
Model & $\tau=0.7$ & $\tau=0.8$ & $\tau=0.9$ \\
\midrule
Qwen3-8B & 78.6 / 5.7k & 81.5 / 6.7k & 83.1 / 7.7k \\
Qwen3-4B & 78.2 / 5.7k & 80.7 / 6.2k & 81.7 / 7.3k \\
R1-Distill-Llama & 50.7 / 6.7k & 52.8 / 7.4k & 54.0 / 8.2k \\
Gemini2.5-Flash & 78.1 / 2.8k & 80.2 / 3.3k & 81.3 / 3.6k \\
\bottomrule
\end{tabular}
\caption{Sensitivity of TRACE to the stopping threshold $\tau$ on overall benchmarks, with fixed $k=5$ and $\alpha=0.7$. Each entry reports accuracy / average tokens.}
\label{tab:tau_sensitivity}
\vspace{-1.2em}
\end{table}

\subsection{Overhead of Answer Induction}
TRACE introduces a lightweight auxiliary decoding step for answer induction. As shown in Table~\ref{tab:induction_overhead}, induction tokens account for only 2--3\% of total reasoning tokens across models (2.23\% on average). For instance, on Qwen3-8B, only 167 out of 8088 tokens (2.06\%) are used for induction. 
This overhead is minimal since induction generates only a short answer and reuses the KV cache. Consequently, TRACE still achieves a clear net reduction in total decoding cost compared to full-length reasoning.

\begin{table}[t]
\centering
\small
\setlength{\tabcolsep}{4pt}
\begin{tabular}{lccc}
\toprule
Model & Induction & Total & Ratio \\
\midrule
Qwen3-8B & 167 & 8088 & 2.06\% \\
Qwen3-4B & 198 & 9489 & 2.08\% \\
Gemini-2.5 & 120 & 4291 & 2.79\% \\
R1-Distill-Llama & 111 & 4784 & 2.32\% \\
\midrule
Average & 149 & 6663 & 2.23\% \\
\bottomrule
\end{tabular}
\caption{Overhead of answer induction in TRACE. Induction tokens account for only a small fraction of total reasoning tokens across models.}
\vspace{-1.2em}
\label{tab:induction_overhead}
\end{table}

\section{Related Work}
\paragraph{Confidence Calibration in LLMs.}
Confidence calibration aims to align a model’s reported confidence with the empirical likelihood that its predictions are correct ~\citep{geng2024survey_llm_calibration, DBLP:conf/kdd/LiuCDC0025}.
Recent evidence further shows that extended reasoning can impair calibration by amplifying confidence in incorrect intermediate hypotheses~\citep{lacombe2025overreasoning_impairs_calibration}.
Prior work has explored this issue through fidelity-based confidence estimators~\citep{zhang2024calibrating_llm_confidence, DBLP:conf/aaai/Dong0PY25}, post-hoc calibration frameworks ~\citep{DBLP:conf/iclr/ManggalaMKKR25} and prompting strategies that encourage more faithful uncertainty expression~\citep{zhao2024far_calibration,DBLP:conf/acl/FloresEC25}.
These findings motivate our approach, which improves confidence calibration by aggregating confidence signals across multiple steps.

\paragraph{Adaptive Computation in LLMs.}
Methods for mitigating overthinking broadly fall into training-free and training-based approaches.
Training-free methods adapt inference using behavioral signals such as entropy
~\citep{chen2025rstitchdynamictrajectorystitching}
or confidence estimates
~\citep{yang2025dynamic},
enabling early exit, model switching, or discourse suppression
~\citep{yang2025dynamic, chen2025rstitchdynamictrajectorystitching, wang-etal-2025-wait}.
Training-based approaches modify models via reinforcement learning with length-aware objectives
~\citep{kimiteam2025kimik15scalingreinforcement, arora2025training}
or fine-tuning on concise reasoning traces
~\citep{yu2025z1efficienttesttimescaling}.
Unlike prior approaches, our method aggregates evidence across multiple steps to achieve more reliable early exit and improved accuracy--efficiency trade-off.

\section{Conclusion}
We present \textbf{TRACE}, a multi-step early-exit framework for efficient and reliable reasoning.
TRACE decides when to stop by detecting reasoning convergence through evidence aggregated across multiple steps, rather than relying on single-step confidence.
It combines two complementary signals: answer consistency (stability of predicted answers) and confidence trajectory (multi-step confidence support).
Across diverse backbone models and math reasoning benchmarks, TRACE maintains accuracy while substantially reducing inference cost, demonstrating robust and general early-exit behavior.

\section*{Limitations}
Our experiments focus on text-only mathematical reasoning benchmarks and a limited set of backbone language models.
While TRACE is designed as a general early-exit criterion based on multi-step convergence signals, we do not evaluate it in multimodal settings (e.g., vision-language reasoning) where intermediate representations and uncertainty may behave differently.
Extending TRACE to multimodal models and tasks, and validating its effectiveness under multimodal reasoning traces, is an important direction for future work.

\bibliography{custom}

\begin{thebibliography}{39}
\providecommand{\natexlab}[1]{#1}

\bibitem[{{AI-MO Project} and {Project Numina}(2025)}]{aimo_validation_amc}
{AI-MO Project} and {Project Numina}. 2025.
\newblock Ai-mo/aimo-validation-amc: {AMC} math validation set.
\newblock \url{https://huggingface.co/datasets/AI-MO/aimo-validation-amc}.
\newblock Dataset on Hugging Face. Extracted from AMC12 2022 and AMC12 2023 problems from the Art of Problem Solving Wiki and adapted to integer outputs.

\bibitem[{Arora and Zanette(2025)}]{arora2025training}
Daman Arora and Andrea Zanette. 2025.
\newblock \href {https://doi.org/10.48550/ARXIV.2502.04463} {Training language models to reason efficiently}.
\newblock \emph{CoRR}, abs/2502.04463.

\bibitem[{Chen et~al.(2025{\natexlab{a}})Chen, Xu, Liang, He, Pang, Yu, Song, Liu, Zhou, Zhang, Wang, Tu, Mi, and Yu}]{chen2024not}
Xingyu Chen, Jiahao Xu, Tian Liang, Zhiwei He, Jianhui Pang, Dian Yu, Linfeng Song, Qiuzhi Liu, Mengfei Zhou, Zhuosheng Zhang, Rui Wang, Zhaopeng Tu, Haitao Mi, and Dong Yu. 2025{\natexlab{a}}.
\newblock \href {https://openreview.net/forum?id=MSbU3L7V00} {Do {NOT} think that much for 2+3=? on the overthinking of long reasoning models}.
\newblock In \emph{ICML 2025}. OpenReview.net.

\bibitem[{Chen et~al.(2025{\natexlab{b}})Chen, Chen, He, Tan, Cai, and Zhuang}]{chen2025rstitchdynamictrajectorystitching}
Zhuokun Chen, Zeren Chen, Jiahao He, Mingkui Tan, Jianfei Cai, and Bohan Zhuang. 2025{\natexlab{b}}.
\newblock \href {https://doi.org/10.48550/ARXIV.2507.17307} {R-stitch: Dynamic trajectory stitching for efficient reasoning}.
\newblock \emph{CoRR}, abs/2507.17307.

\bibitem[{Committees(2024)}]{aime_problems_solutions}
MAA Committees. 2024.
\newblock \href {https://artofproblemsolving.com/wiki/index.php/AIME_Problems_and_Solutions} {Aime problems and solutions}.
\newblock Online.
\newblock Retrieved from Art of Problem Solving Wiki.

\bibitem[{Cuadron et~al.(2025)Cuadron, Li, Ma, Wang, Wang, Zhuang, Liu, Schroeder, Xia, Mao, Thumiger, Desai, Stoica, Klimovic, Neubig, and Gonzalez}]{cuadron2025danger}
Alejandro Cuadron, Dacheng Li, Wenjie Ma, Xingyao Wang, Yichuan Wang, Siyuan Zhuang, Shu Liu, Luis~Gaspar Schroeder, Tian Xia, Huanzhi Mao, Nicholas Thumiger, Aditya Desai, Ion Stoica, Ana Klimovic, Graham Neubig, and Joseph~E. Gonzalez. 2025.
\newblock \href {https://doi.org/10.48550/ARXIV.2502.08235} {The danger of overthinking: Examining the reasoning-action dilemma in agentic tasks}.
\newblock \emph{CoRR}, abs/2502.08235.

\bibitem[{DeepSeek{-}AI(2025)}]{guo2025deepseek}
DeepSeek{-}AI. 2025.
\newblock \href {https://doi.org/10.48550/ARXIV.2501.12948} {Deepseek-r1: Incentivizing reasoning capability in llms via reinforcement learning}.
\newblock \emph{CoRR}, abs/2501.12948.

\bibitem[{Dong et~al.(2025)Dong, Jiang, Pan, and Yu}]{DBLP:conf/aaai/Dong0PY25}
Jinzong Dong, Zhaohui Jiang, Dong Pan, and Haoyang Yu. 2025.
\newblock \href {https://doi.org/10.1609/AAAI.V39I15.33792} {Combining priors with experience: Confidence calibration based on binomial process modeling}.
\newblock In \emph{AAAI 25}, pages 16317--16326. {AAAI} Press.

\bibitem[{Flores et~al.(2025)Flores, Ernst, and Cheung}]{DBLP:conf/acl/FloresEC25}
Lorenzo Jaime~Yu Flores, Ori Ernst, and Jackie~CK Cheung. 2025.
\newblock \href {https://doi.org/10.18653/V1/2025.ACL-SHORT.15} {Improving the calibration of confidence scores in text generation using the output distribution's characteristics}.
\newblock In \emph{ACL 2025}, pages 172--182. Association for Computational Linguistics.

\bibitem[{Fu et~al.(2024)Fu, Chen, Zhu, Fu, Dai, Qiao, and Zhang}]{fu2024efficiently}
Yichao Fu, Junda Chen, Siqi Zhu, Zheyu Fu, Zhongdongming Dai, Aurick Qiao, and Hao Zhang. 2024.
\newblock \href {https://doi.org/10.48550/ARXIV.2412.20993} {Efficiently serving {LLM} reasoning programs with certaindex}.
\newblock \emph{CoRR}, abs/2412.20993.

\bibitem[{Geng et~al.(2024)Geng, Cai, Wang, Koeppl, Nakov, and Gurevych}]{geng2024survey_llm_calibration}
Jiahui Geng, Fengyu Cai, Yuxia Wang, Heinz Koeppl, Preslav Nakov, and Iryna Gurevych. 2024.
\newblock \href {https://doi.org/10.18653/v1/2024.naacl-long.366} {A survey of confidence estimation and calibration in large language models}.
\newblock In \emph{NAACL 2024}, pages 6577--6595. Association for Computational Linguistics.

\bibitem[{Guan et~al.(2025)Guan, Zhang, Liu, Shang, Sun, Zhu, Yang, and Yang}]{guan2025rstar}
Xinyu Guan, Li~Lyna Zhang, Yifei Liu, Ning Shang, Youran Sun, Yi~Zhu, Fan Yang, and Mao Yang. 2025.
\newblock \href {https://openreview.net/forum?id=5zwF1GizFa} {rstar-math: Small llms can master math reasoning with self-evolved deep thinking}.
\newblock In \emph{ICML 2025}. OpenReview.net.

\bibitem[{Han et~al.(2025)Han, Wang, Fang, Zhao, Ma, and Chen}]{han2025token}
Tingxu Han, Zhenting Wang, Chunrong Fang, Shiyu Zhao, Shiqing Ma, and Zhenyu Chen. 2025.
\newblock \href {https://aclanthology.org/2025.findings-acl.1274/} {Token-budget-aware {LLM} reasoning}.
\newblock In \emph{ACL 2025}, pages 24842--24855. Association for Computational Linguistics.

\bibitem[{He et~al.(2024)He, Luo, Bai, Hu, Thai, Shen, Hu, Han, Huang, Zhang, Liu, Qi, Liu, and Sun}]{he2024olympiadbench}
Chaoqun He, Renjie Luo, Yuzhuo Bai, Shengding Hu, Zhen~Leng Thai, Junhao Shen, Jinyi Hu, Xu~Han, Yujie Huang, Yuxiang Zhang, Jie Liu, Lei Qi, Zhiyuan Liu, and Maosong Sun. 2024.
\newblock \href {https://doi.org/10.18653/V1/2024.ACL-LONG.211} {Olympiadbench: {A} challenging benchmark for promoting {AGI} with olympiad-level bilingual multimodal scientific problems}.
\newblock In \emph{ACL 2024}, pages 3828--3850. Association for Computational Linguistics.

\bibitem[{Huang et~al.(2025)Huang, Lin, Feng, Chen, He, and Hou}]{huang2025efficient}
Jiameng Huang, Baijiong Lin, Guhao Feng, Jierun Chen, Di~He, and Lu~Hou. 2025.
\newblock \href {https://doi.org/10.48550/ARXIV.2508.05337} {Efficient reasoning for large reasoning language models via certainty-guided reflection suppression}.
\newblock \emph{CoRR}, abs/2508.05337.

\bibitem[{Lacombe et~al.(2025)Lacombe, Wu, and Dilworth}]{lacombe2025overreasoning_impairs_calibration}
Romain Lacombe, Kerrie Wu, and Eddie Dilworth. 2025.
\newblock \href {https://doi.org/10.48550/ARXIV.2508.15050} {Don't think twice! over-reasoning impairs confidence calibration}.
\newblock \emph{CoRR}, abs/2508.15050.

\bibitem[{Li et~al.(2025)Li, Yuan, Yu, Guo, and Cao}]{li2025cocoevo}
Kefan Li, Yuan Yuan, Hongyue Yu, Tingyu Guo, and Shijie Cao. 2025.
\newblock \href {https://doi.org/10.48550/ARXIV.2502.10802} {Cocoevo: Co-evolution of programs and test cases to enhance code generation}.
\newblock \emph{TEVC 2025}.

\bibitem[{Lightman et~al.(2024)Lightman, Kosaraju, Burda, Edwards, Baker, Lee, Leike, Schulman, Sutskever, and Cobbe}]{lightman2023let}
Hunter Lightman, Vineet Kosaraju, Yuri Burda, Harrison Edwards, Bowen Baker, Teddy Lee, Jan Leike, John Schulman, Ilya Sutskever, and Karl Cobbe. 2024.
\newblock \href {https://openreview.net/forum?id=v8L0pN6EOi} {Let's verify step by step}.
\newblock In \emph{ICLR 2024}. OpenReview.net.

\bibitem[{Liu et~al.(2025)Liu, Chen, Da, Chen, Lin, and Wei}]{DBLP:conf/kdd/LiuCDC0025}
Xiaoou Liu, Tiejin Chen, Longchao Da, Chacha Chen, Zhen Lin, and Hua Wei. 2025.
\newblock \href {https://doi.org/10.1145/3711896.3736569} {Uncertainty quantification and confidence calibration in large language models: {A} survey}.
\newblock In \emph{KDD 2025}, pages 6107--6117. {ACM}.

\bibitem[{Liu and Wang(2025)}]{liu2025answerconvergencesignalearly}
Xin Liu and Lu~Wang. 2025.
\newblock \href {https://doi.org/10.48550/ARXIV.2506.02536} {Answer convergence as a signal for early stopping in reasoning}.
\newblock \emph{CoRR}, arXiv:2506.02536.

\bibitem[{Lu et~al.(2025)Lu, Han, Acuna, Kim, Jung, Prabhumoye, Muennighoff, Patwary, Shoeybi, Catanzaro, and Choi}]{lu2025retro}
Ximing Lu, Seungju Han, David Acuna, Hyunwoo Kim, Jaehun Jung, Shrimai Prabhumoye, Niklas Muennighoff, Mostofa Patwary, Mohammad Shoeybi, Bryan Catanzaro, and Yejin Choi. 2025.
\newblock \href {https://doi.org/10.48550/ARXIV.2504.04383} {Retro-search: Exploring untaken paths for deeper and efficient reasoning}.
\newblock \emph{CoRR}, abs/2504.04383.

\bibitem[{Manggala et~al.(2025)Manggala, Mastakouri, Kirschbaum, Kasiviswanathan, and Ramdas}]{DBLP:conf/iclr/ManggalaMKKR25}
Putra Manggala, Atalanti{-}Anastasia Mastakouri, Elke Kirschbaum, Shiva~Prasad Kasiviswanathan, and Aaditya Ramdas. 2025.
\newblock \href {https://openreview.net/forum?id=D2hhkU5O48} {Qa-calibration of language model confidence scores}.
\newblock In \emph{ICLR 2025}. OpenReview.net.

\bibitem[{Rein et~al.(2023)Rein, Hou, Stickland, Petty, Pang, Dirani, Michael, and Bowman}]{rein2024gpqa}
David Rein, Betty~Li Hou, Asa~Cooper Stickland, Jackson Petty, Richard~Yuanzhe Pang, Julien Dirani, Julian Michael, and Samuel~R. Bowman. 2023.
\newblock \href {https://doi.org/10.48550/ARXIV.2311.12022} {{GPQA:} {A} graduate-level google-proof q{\&}a benchmark}.
\newblock \emph{CoRR}, abs/2311.12022.

\bibitem[{Snell et~al.(2024)Snell, Lee, Xu, and Kumar}]{snell2024scalingllmtesttimecompute}
Charlie Snell, Jaehoon Lee, Kelvin Xu, and Aviral Kumar. 2024.
\newblock \href {https://doi.org/10.48550/ARXIV.2408.03314} {Scaling {LLM} test-time compute optimally can be more effective than scaling model parameters}.
\newblock \emph{CoRR}, arXiv:2408.03314.

\bibitem[{Sui et~al.(2025)Sui, Chuang, Wang, Zhang, Zhang, Yuan, Liu, Wen, Zhong, Zou, Chen, and Hu}]{sui2025stop}
Yang Sui, Yu{-}Neng Chuang, Guanchu Wang, Jiamu Zhang, Tianyi Zhang, Jiayi Yuan, Hongyi Liu, Andrew Wen, Shaochen Zhong, Na~Zou, Hanjie Chen, and Xia Hu. 2025.
\newblock \href {https://openreview.net/forum?id=HvoG8SxggZ} {Stop overthinking: {A} survey on efficient reasoning for large language models}.
\newblock \emph{Trans. Mach. Learn. Res.}, 2025.

\bibitem[{Team(2025)}]{comanici2025gemini}
Gemini Team. 2025.
\newblock \href {https://doi.org/10.48550/ARXIV.2507.06261} {Gemini 2.5: Pushing the frontier with advanced reasoning, multimodality, long context, and next generation agentic capabilities}.
\newblock \emph{CoRR}, abs/2507.06261.

\bibitem[{Team et~al.(2025)Team, Du, Gao, Xing, Jiang, Chen, Li, Xiao, Du, Liao, Tang, Wang, Zhang, Yuan, Lu, Tang, Sung, Wei, Lai, Guo, Zhu, Ding, Hu, Yang, Zhang, Yao, Zhao, Lu, Li, Yu, Gao, Zheng, Yuan, Chen, Guo, Su, Wang, Zhao, Zhang, Liu, Yan, Wu, Shi, Ye, Yu, Dong, Zhang, Ma, Pan, Gong, Liu, Ma, Wei, Cao, Huang, Jiang, Gao, Xiong, He, Huang, Wu, He, Wei, Jia, Wu, Xu, Zu, Zhou, Pan, Charles, Li, Hu, Liu, Chen, Wang, Liu, Qin, Liu, Yang, Bao, Du, Wu, Wang, Zhou, Wang, Li, Zhu, Zhang, Wang, Yang, Huang, Huang, Xu, and Yang}]{kimiteam2025kimik15scalingreinforcement}
Kimi Team, Angang Du, Bofei Gao, Bowei Xing, Changjiu Jiang, Cheng Chen, Cheng Li, Chenjun Xiao, Chenzhuang Du, Chonghua Liao, Chuning Tang, Congcong Wang, Dehao Zhang, Enming Yuan, Enzhe Lu, Fengxiang Tang, Flood Sung, Guangda Wei, Guokun Lai, and 75 others. 2025.
\newblock \href {https://doi.org/10.48550/ARXIV.2501.12599} {Kimi k1.5: Scaling reinforcement learning with llms}.
\newblock \emph{CoRR}, arXiv:2501.12599.

\bibitem[{Wang et~al.(2025)Wang, Feng, Chen, Chu, Krishna, and Zhou}]{wang-etal-2025-wait}
Chenlong Wang, Yuanning Feng, Dongping Chen, Zhaoyang Chu, Ranjay Krishna, and Tianyi Zhou. 2025.
\newblock \href {https://doi.org/10.18653/v1/2025.findings-emnlp.394} {Wait, we don{'}t need to ``wait''! removing thinking tokens improves reasoning efficiency}.
\newblock In \emph{EMNLP 2025}, pages 7459--7482. Association for Computational Linguistics.

\bibitem[{Wang et~al.(2023)Wang, Wei, Schuurmans, Le, Chi, Narang, Chowdhery, and Zhou}]{wang2023selfconsistencyimproveschainthought}
Xuezhi Wang, Jason Wei, Dale Schuurmans, Quoc~V. Le, Ed~H. Chi, Sharan Narang, Aakanksha Chowdhery, and Denny Zhou. 2023.
\newblock \href {https://openreview.net/forum?id=1PL1NIMMrw} {Self-consistency improves chain of thought reasoning in language models}.
\newblock In \emph{ICLR 2023}. OpenReview.net.

\bibitem[{Wei et~al.(2022)Wei, Wang, Schuurmans, Bosma, Ichter, Xia, Chi, Le, and Zhou}]{wei2023chainofthoughtpromptingelicitsreasoning}
Jason Wei, Xuezhi Wang, Dale Schuurmans, Maarten Bosma, Brian Ichter, Fei Xia, Ed~H. Chi, Quoc~V. Le, and Denny Zhou. 2022.
\newblock \href {http://papers.nips.cc/paper\_files/paper/2022/hash/9d5609613524ecf4f15af0f7b31abca4-Abstract-Conference.html} {Chain-of-thought prompting elicits reasoning in large language models}.
\newblock In \emph{NeurIPS 2022}.

\bibitem[{Xiong et~al.(2024)Xiong, Hu, Lu, Li, Fu, He, and Hooi}]{xiong2023can}
Miao Xiong, Zhiyuan Hu, Xinyang Lu, Yifei Li, Jie Fu, Junxian He, and Bryan Hooi. 2024.
\newblock \href {https://openreview.net/forum?id=gjeQKFxFpZ} {Can llms express their uncertainty? an empirical evaluation of confidence elicitation in llms}.
\newblock In \emph{ICLR 2024}. OpenReview.net.

\bibitem[{Xu et~al.(2025)Xu, Hao, Zong, Wang, Zhang, Wang, Lan, Gong, Ouyang, Meng, Shao, Yan, Yang, Song, Ren, Hu, Li, Feng, Gao, and Li}]{xu2025largereasoningmodelssurvey}
Fengli Xu, Qianyue Hao, Zefang Zong, Jingwei Wang, Yunke Zhang, Jingyi Wang, Xiaochong Lan, Jiahui Gong, Tianjian Ouyang, Fanjin Meng, Chenyang Shao, Yuwei Yan, Qinglong Yang, Yiwen Song, Sijian Ren, Xinyuan Hu, Yu~Li, Jie Feng, Chen Gao, and Yong Li. 2025.
\newblock \href {https://doi.org/10.48550/ARXIV.2501.09686} {Towards large reasoning models: {A} survey of reinforced reasoning with large language models}.
\newblock \emph{CoRR}, arXiv:2501.09686.

\bibitem[{Yang et~al.(2025{\natexlab{a}})Yang, Li, Yang, Zhang, Hui, Zheng, Yu, Gao, Huang, Lv, Zheng, Liu, Zhou, Huang, Hu, Ge, Wei, Lin, Tang, Yang, Tu, Zhang, Yang, Yang, Zhou, Lin, Dang, Bao, Yang, Yu, Deng, Li, Xue, Li, Zhang, Wang, Zhu, Men, Gao, Liu, Luo, Li, Tang, Yin, Ren, Wang, Zhang, Ren, Fan, Su, Zhang, Zhang, Wan, Liu, Wang, Cui, Zhang, Zhou, and Qiu}]{yang2025qwen3technicalreport}
An~Yang, Anfeng Li, Baosong Yang, Beichen Zhang, Binyuan Hui, Bo~Zheng, Bowen Yu, Chang Gao, Chengen Huang, Chenxu Lv, Chujie Zheng, Dayiheng Liu, Fan Zhou, Fei Huang, Feng Hu, Hao Ge, Haoran Wei, Huan Lin, Jialong Tang, and 40 others. 2025{\natexlab{a}}.
\newblock \href {https://doi.org/10.48550/ARXIV.2505.09388} {Qwen3 technical report}.
\newblock \emph{CoRR}, arXiv:2505.09388.

\bibitem[{Yang et~al.(2025{\natexlab{b}})Yang, Si, Duan, Zhu, Zhu, Lin, Cao, and Wang}]{yang2025dynamic}
Chenxu Yang, Qingyi Si, Yongjie Duan, Zheliang Zhu, Chenyu Zhu, Zheng Lin, Li~Cao, and Weiping Wang. 2025{\natexlab{b}}.
\newblock \href {https://doi.org/10.48550/ARXIV.2504.15895} {Dynamic early exit in reasoning models}.
\newblock \emph{CoRR}, abs/2504.15895.

\bibitem[{Yang et~al.(2025{\natexlab{c}})Yang, Ma, Lin, and Wei}]{yang2025towards}
Wenkai Yang, Shuming Ma, Yankai Lin, and Furu Wei. 2025{\natexlab{c}}.
\newblock \href {https://doi.org/10.48550/ARXIV.2502.18080} {Towards thinking-optimal scaling of test-time compute for {LLM} reasoning}.
\newblock \emph{CoRR}, abs/2502.18080.

\bibitem[{Yao et~al.(2023)Yao, Yu, Zhao, Shafran, Griffiths, Cao, and Narasimhan}]{yao2023tree}
Shunyu Yao, Dian Yu, Jeffrey Zhao, Izhak Shafran, Tom Griffiths, Yuan Cao, and Karthik Narasimhan. 2023.
\newblock \href {http://papers.nips.cc/paper\_files/paper/2023/hash/271db9922b8d1f4dd7aaef84ed5ac703-Abstract-Conference.html} {Tree of thoughts: Deliberate problem solving with large language models}.
\newblock In \emph{NeurIPS 2023}.

\bibitem[{Yu et~al.(2025)Yu, Wu, Zhao, Cohan, and Zhang}]{yu2025z1efficienttesttimescaling}
Zhaojian Yu, Yinghao Wu, Yilun Zhao, Arman Cohan, and Xiao{-}Ping Zhang. 2025.
\newblock \href {https://doi.org/10.48550/ARXIV.2504.00810} {{Z1:} efficient test-time scaling with code}.
\newblock \emph{CoRR}, arXiv:2504.00810.

\bibitem[{Zhang et~al.(2024)Zhang, Huang, Shi, Guo, Peng, Yan, Zhou, and Qiu}]{zhang2024calibrating_llm_confidence}
Mozhi Zhang, Mianqiu Huang, Rundong Shi, Linsen Guo, Chong Peng, Peng Yan, Yaqian Zhou, and Xipeng Qiu. 2024.
\newblock \href {https://doi.org/10.18653/V1/2024.EMNLP-MAIN.173} {Calibrating the confidence of large language models by eliciting fidelity}.
\newblock In \emph{EMNLP 2024}, pages 2959--2979. Association for Computational Linguistics.

\bibitem[{Zhao et~al.(2024)Zhao, Zhang, Pan, Yao, Yu, Wu, and Chen}]{zhao2024far_calibration}
Xinran Zhao, Hongming Zhang, Xiaoman Pan, Wenlin Yao, Dong Yu, Tongshuang Wu, and Jianshu Chen. 2024.
\newblock \href {https://doi.org/10.18653/V1/2024.FINDINGS-ACL.515} {Fact-and-reflection (far) improves confidence calibration of large language models}.
\newblock In \emph{ACL 2024}, pages 8702--8718. Association for Computational Linguistics.

\end{thebibliography}

\appendix

\section{Experimental Settings}
\subsection{Hyper-parameters Setting}
\label{app:Hyper-parameters Setting}
We detail the hyperparameter choices used in TRACE for computing answer consistency and termination scores. The early-exit threshold and window size are selected on held-out validation sets and fixed across all benchmarks.

For the Answer Consistency Score (ACS) defined in Eq.~(\ref{eq:acs_score}), consistency is computed over a sliding window of the most recent \(k\) induced answers. In our experiments, we set the window size to \(k =  5\), which balances robustness to transient fluctuations with responsiveness to answer changes.

For the combined termination score \(S(a)\) defined in Eq.~(\ref{eq:sa_caculate}), we compute a weighted combination of ACS and the Confidence Trajectory Score (CTS). The weighting coefficient \(\alpha\) is set to either \(0.7\) or \(0.3\), controlling the relative contribution of answer stability and confidence dynamics.

Finally, reasoning is terminated when the combined score exceeds a predefined threshold. Unless otherwise specified, we set the termination threshold to \(0.8\) across all experiments.

\subsection{Datasets Details}
The mathematical and scientific reasoning datasets used in our evaluation are described below.
We strictly follow the licenses specified in the original papers.

OlympiadBench \citep{he2024olympiadbench} is a challenging benchmark designed to evaluate advanced mathematical reasoning at the Olympiad level.
It consists of problems sourced from international and national mathematics competitions, covering topics such as algebra, geometry, number theory, and combinatorics.
The problems typically require long and structured multi-step reasoning, often involving symbolic manipulation and proof-like reasoning processes, making the benchmark particularly difficult for LLMs.

MATH500 \citep{lightman2023let} is a curated subset of the MATH dataset, containing 500 high-difficulty competition-style math problems.
The problems span multiple domains, including algebra, calculus, geometry, and probability, and require explicit chain-of-thought reasoning to arrive at the final answer.
MATH500 is commonly used as a standardized test set for evaluating mathematical reasoning performance under zero-shot or test-time scaling settings.

AIME24 and AIME25 \citep{aime_problems_solutions} are drawn from the American Invitational Mathematics Examination (AIME) problems from 2024 and 2025, respectively.
Each dataset consists of short-answer competition problems.
These problems often involve deep multi-step reasoning and nontrivial mathematical insights, and are widely regarded as a strong test of symbolic and numerical reasoning capabilities.

AMC23 \citep{aimo_validation_amc} is a validation benchmark constructed from AMC 12 problems released in 2023.
The dataset focuses on high-school–level mathematics and includes problems requiring multiple reasoning steps, such as algebraic transformations, geometric reasoning, and combinatorial counting.
Following prior work, all answers are normalized to integer outputs for consistent evaluation.

GPQA-D \citep{rein2024gpqa} is the difficult split of the Graduate-Level Google-Proof QA (GPQA) benchmark, designed to assess scientific reasoning beyond mathematics.
It consists of expert-written multiple-choice questions in physics, chemistry, and biology, where correct answers require domain knowledge and multi-step logical reasoning rather than surface-level pattern matching.
GPQA-D is considered particularly challenging due to its resistance to memorization and retrieval-based shortcuts.

All datasets are evaluated in a zero-shot inference setting.
Their inherent requirement for long reasoning chains and complex intermediate computations makes them well-suited for analyzing early-exit strategies and test-time scaling behavior.
\label{app:datasets_details}

\subsection{Prompt Templates}
\label{app:prompt_templates}

To ensure fair and consistent evaluation across model families, we use model-specific prompt templates that match each model's expected input format (chat-style vs.\ plain-text) while keeping the task instruction semantically identical. In all cases, we explicitly require the model to place the final answer in \texttt{\textbackslash boxed\{\}} so that downstream answer parsing is standardized across models. Table~\ref{tab:prompt_templates} summarizes the simplified templates used for Qwen, R1-Distill-LLaMA, and Gemini-2.5-Flash.

\label{app:prompt_templates}

\begin{table}[t]
\centering
\small
\setlength{\tabcolsep}{6pt}
\begin{tabular}{p{0.26\linewidth} p{0.68\linewidth}}
\hline
\textbf{Model family} & \textbf{Prompt template (simplified)} \\
\hline

Qwen3 (Qwen-series) &
Chat-format prompt with the instruction:
\emph{``Please reason step by step, and put your final answer within \texttt{\textbackslash boxed\{\}}.''}
We optionally prepend a thinking tag (e.g., \texttt{<think>}) when supported. \\

R1-Distill-LLaMA &
Chat-format prompt with expert-style instruction:
\emph{``Think about the problem internally. Then output a short explanation, and put your final answer within \texttt{\textbackslash boxed\{\}}.''} \\

Gemini2.5-Flash &
Plain-text instruction prompt:
\emph{``You are a careful and logical assistant. Please reason step by step, and put your final answer within \texttt{\textbackslash boxed\{\}}.''}
followed by \emph{Question: \{question\}} and \emph{Solution:}. \\

\hline
\end{tabular}
\caption{Prompt templates used for different model families. We standardize outputs by requiring the final answer to appear in \texttt{\textbackslash boxed\{\}} for consistent parsing and evaluation.}
\vspace{-1.5em}
\label{tab:prompt_templates}
\end{table}

\section{Implementation Details of TRACE}
\label{app:trace_impl}

This appendix provides implementation details of TRACE, including the step-wise answer induction, confidence computation, sliding-window aggregation, and the early-exit decision logic.

\subsection{Overview}
At inference time, TRACE monitors a multi-step reasoning process and decides whether to terminate after each step.
Given the reasoning generated up to step \(t\), TRACE (i) induces a candidate final answer \(a_t\) using a lightweight auxiliary prompt (Appendix~\ref{app:answer_induction}), (ii) computes a scalar confidence \(c_t\) for \(a_t\) from token-level probabilities, and (iii) aggregates evidence within a sliding window of the last \(k\) steps to compute ACS/CTS and the combined stability score in Eq.~(\ref{eq:sa_caculate}).
Inference stops when the best-scoring candidate exceeds the threshold \(\tau\).

\subsection{Pseudo-code}
Algorithm~\ref{alg:trace} summarizes the TRACE procedure.

\begin{algorithm*}[t]
\caption{TRACE: Temporal Reasoning Aggregation for Convergent Exit}
\label{alg:trace}
\begin{algorithmic}[1]
\Require Question \(q\); model \(M\); window size \(k\); weight \(\alpha\); threshold \(\tau\); max steps \(T_{\max}\)
\Ensure Final answer \(a_{\mathrm{final}}\)

\State \(\mathcal{W} \leftarrow \emptyset\) \Comment{Sliding window buffer of tuples \((a_t, c_t)\)}
\State \(R \leftarrow \emptyset\) \Comment{Accumulated reasoning text}

\For{\(t = 1\) \textbf{to} \(T_{\max}\)}
    \State \(r_t \leftarrow M.\textsc{GenerateStep}(q, R)\) \Comment{Generate the next reasoning step}
    \State \(R \leftarrow R \cup r_t\)

    \State \(a_t \leftarrow \textsc{InduceAnswer}(q, R)\) \Comment{Auxiliary prompt; returns an induced answer}
    \State \(c_t \leftarrow \textsc{TokenEntropyConfidence}(a_t)\) \Comment{Eqs.~(\ref{eq:norm_entropy})--(\ref{eq:confidence_caculate})}

    \State Append \((a_t, c_t)\) to \(\mathcal{W}\); if \(|\mathcal{W}| > k\) remove the oldest item
    \State \(\mathcal{A} \leftarrow\) set of unique answers appearing in \(\mathcal{W}\)

    \For{\(\textbf{each } a \in \mathcal{A}\)}
        \State \(\mathrm{ACS}(a) \leftarrow \frac{\mathrm{count}(a \text{ in } \mathcal{W})}{k}\) \Comment{Eq.~(\ref{eq:acs_score})}
        \State \(\mathrm{CTS}(a) \leftarrow \textsc{AvgConf}(a, \mathcal{W})\) \Comment{Eq.~(\ref{eq:cts_score})}
        \State \(S(a) \leftarrow \alpha \cdot \mathrm{ACS}(a) + (1-\alpha)\cdot \mathrm{CTS}(a)\) \Comment{Eq.~(\ref{eq:sa_caculate})}
    \EndFor

    \State \(a^\star \leftarrow \arg\max_{a \in \mathcal{A}} S(a)\) \Comment{Eq.~(\ref{eq:argmax})}
    \If{\(S(a^\star) \ge \tau\)}
        \State \Return \(a^\star\) \Comment{Early exit: reasoning is considered converged}
    \EndIf
\EndFor

\State \Return \(a_{T_{\max}}\) \Comment{Fallback: return last induced answer}
\end{algorithmic}
\end{algorithm*}

\subsection{Reasoning Step Segmentation via Discourse Markers}
\label{app:step_segmentation}

TRACE requires a step-wise reasoning trace. Since different model families format reasoning differently, we implement a lightweight, model-aware step segmentation procedure based on the frequency of step boundary cues (``discourse markers'') observed during generation.

\paragraph{Streaming segmentation.}
We decode the model output in a streaming manner and maintain the growing text buffer. Every \(n\) generated tokens (we use \(n{=}10\) by default), we scan the recent suffix of the buffer for occurrences of a predefined set of boundary strings (stop tokens). We record the absolute character positions of all newly found matches. Once the number of detected matches reaches a preset limit (\texttt{match\_limit}), we truncate the reasoning trace \emph{before} the \(\texttt{match\_limit}\)-th match position and treat the truncated prefix as the completed trace up to the desired number of steps.

Formally, let \(\mathcal{S}=\{s_1,\dots,s_m\}\) denote the stop-token set, and let \(\{(s, p)\}\) be the ordered list of match events (token \(s\) found at position \(p\)) discovered during decoding. When the \(\texttt{match\_limit}\)-th match event occurs at position \(p^\star\), we output the truncated trace \(R = \textsc{prefix}(R_{\mathrm{full}}, p^\star)\).

\paragraph{Model-specific stop tokens.}
We use different boundary cues for different models:
(i) For Gemini-2.5-Flash, we segment steps using paragraph breaks, i.e., the stop token \texttt{\textbackslash n\textbackslash n}, since Gemini tends to separate reasoning chunks by blank lines.
(ii) For other models (e.g., Qwen and R1-Distill-LLaMA), we use a set of discourse markers that commonly indicate shifts, revisions, or alternative reasoning branches:
\{\texttt{Wait}, \texttt{But}, \texttt{Let me think}, \texttt{</think>}, \texttt{Alternatively}\}.
These markers provide robust step boundaries even when the model does not explicitly number steps.

\paragraph{Practical considerations.}
To avoid missing matches that span across scanning boundaries, we rescan from a small overlap region proportional to the maximum stop-token length. We also deduplicate repeated matches at the same position. In rare cases where generation terminates naturally before reaching \texttt{match\_limit}, we keep the full output and proceed with TRACE using the available steps.

\subsection{Answer Induction and Efficiency Considerations}
\label{app:answer_induction}

To compute the Answer Consistency Score (ACS), we induce a candidate final answer at each reasoning step from the partial trace generated so far. Specifically, at step \(t\) we reuse the existing reasoning context and append a fixed induction prompt (``We can get the question’s Final Answer: \texttt{\textbackslash boxed\{\}}''), instructing the model to output a single answer in \texttt{\textbackslash boxed\{\}} format. This yields an induced-answer sequence \(\{a_1,\ldots,a_T\}\).

Final answers are parsed by taking the last \texttt{\textbackslash boxed\{\}} span when present; otherwise, we fall back to the most explicit short answer in the concluding portion of the response. If multiple candidates occur within a step, we select the one closest to the step’s conclusion and matching the expected final-answer format.

Our induction is lightweight and efficient: it requires only a short auxiliary generation conditioned on the already-produced reasoning trace, without re-generating the trace or interrupting the ongoing reasoning process.

\begin{figure}[t]
    \centering
    \includegraphics[width=\columnwidth]{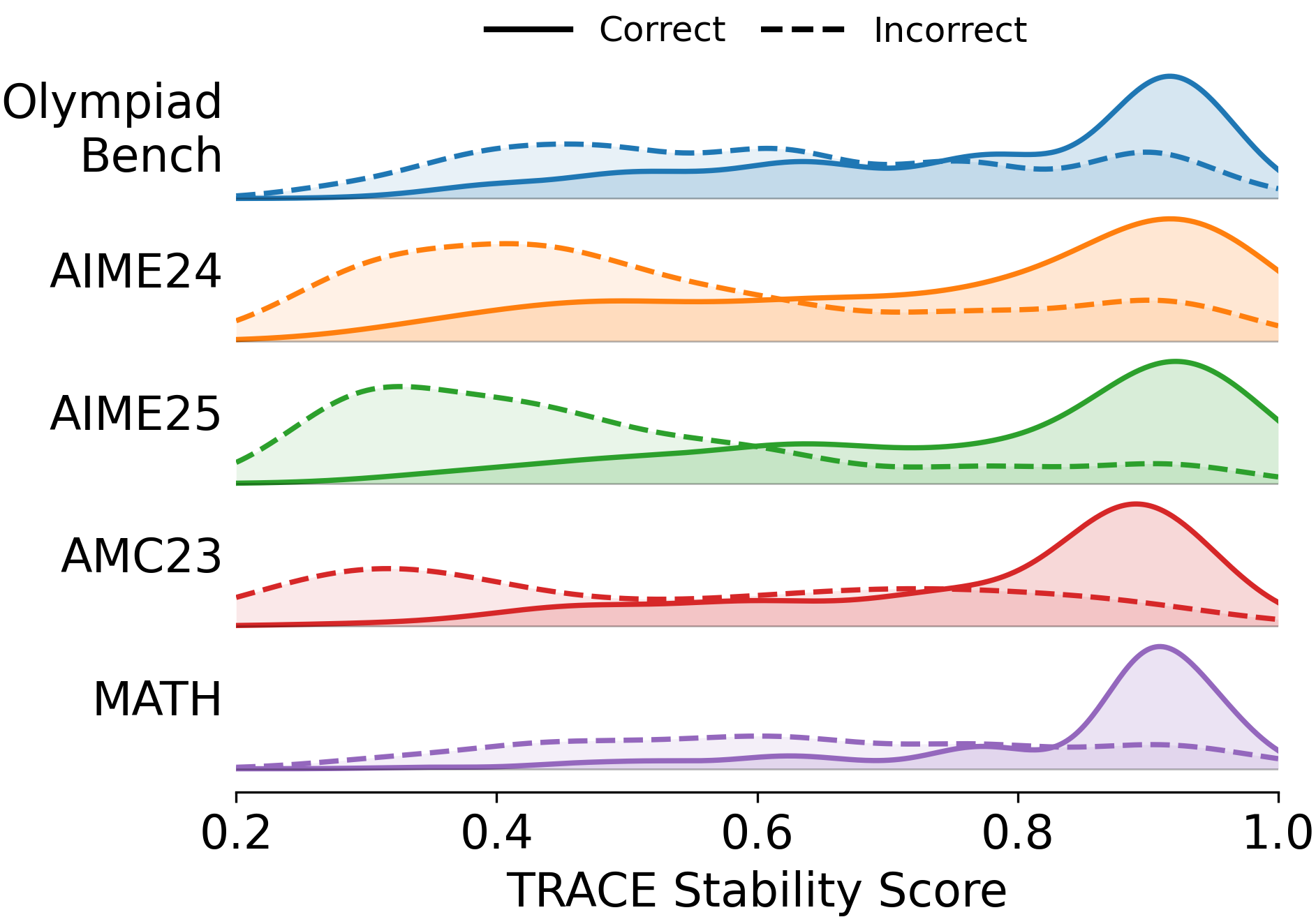}
    \caption{Kernel density estimates of TRACE Stability Score for correct (solid) and incorrect (dashed) predictions across multiple math benchmarks.}
    \label{fig:trace_kde}
\end{figure}

\begin{figure*}[t]
  \centering
  \includegraphics[width=0.7\textwidth]{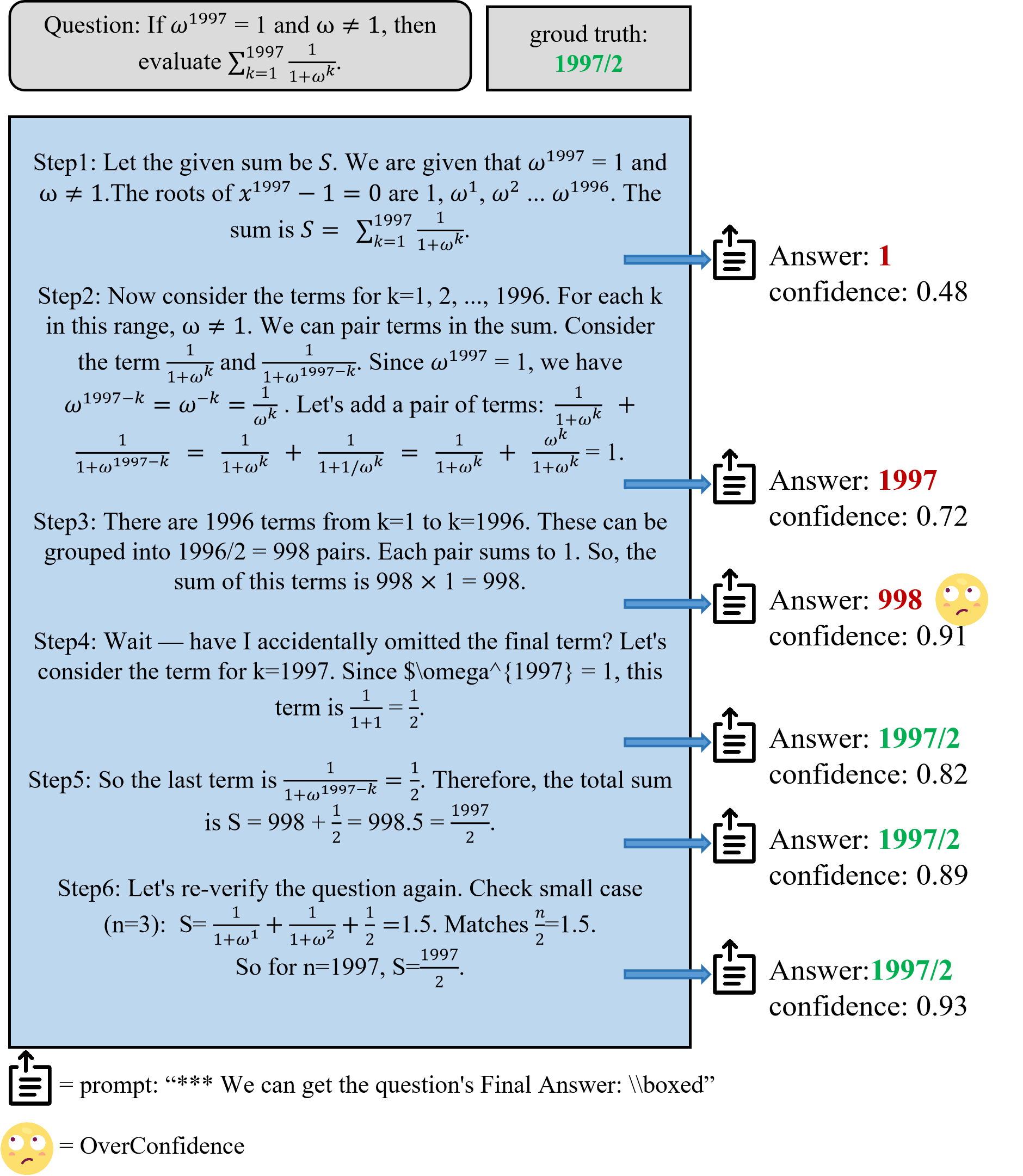}
  \caption{Illustration of a case study.}
  \label{fig:extract_answer}
\end{figure*}

\section{Case Study: Transient Overconfidence vs.\ Multi-step Convergence}
\label{sec:case_study}

Figure~\ref{fig:extract_answer} presents a representative example illustrating why single-step confidence can be unreliable for early termination and how TRACE yields a safer stopping decision by aggregating evidence across multiple steps.
The task is to evaluate a complex-valued summation with ground-truth answer \(1997/2\).
We apply our step-wise answer induction procedure to extract an induced answer and a confidence estimate after each reasoning step.

\paragraph{Failure mode of single-step confidence.}
The model's induced answers fluctuate substantially across steps.
Early in the reasoning, the model proposes several incorrect candidates (e.g., \(1\), \(1997\)), and then produces the answer \(998\) at Step 3 with a \emph{high} confidence of \(0.91\) (marked as overconfident in the figure).
A termination policy based on single-step confidence would likely stop at this point because the confidence is high and appears decisive.
However, this confidence spike is transient and misleading: the model has not fully accounted for the missing final term, which it only realizes later (Step 4).
This example highlights a common overthinking/early-exit pitfall: local confidence can be inflated even when the reasoning has not converged, leading to premature termination with an incorrect answer.

\paragraph{Why TRACE is safer: multi-step evidence aggregation.}
TRACE explicitly avoids making stopping decisions from a single snapshot.
Instead, it tracks whether the reasoning process has \emph{converged} by aggregating signals over recent steps.
In this example, the high-confidence but incorrect answer \(998\) is not stable: it appears only briefly and is preceded by different answers.
Consequently, its Answer Consistency Score (ACS) over a sliding window remains low, preventing early termination even when its single-step confidence is high.
After Step 4, the model corrects the omission and the induced answer becomes \(1997/2\), which then persists in subsequent steps with consistently high confidence.
At this stage, both ACS (answer stability across steps) and CTS (confidence trend/level over recent steps) increase, indicating genuine convergence.
Therefore, TRACE triggers early exit only after the answer stabilizes, reducing susceptibility to transient overconfidence while preserving reasoning capability.

\paragraph{Takeaway.}
This case study demonstrates that single-step confidence can exhibit transient overconfidence and fail to reflect true convergence.
By requiring both (i) repeated agreement of induced answers across multiple steps and (ii) a consistent confidence trajectory, TRACE yields more reliable termination decisions than single-step baselines.

\section{Additional Analysis: Stability Score Distribution}
\label{app:stability_kde}
To further understand the behavior of TRACE, we analyze the distribution of the Stability Score under correct and incorrect predictions.
Figure~\ref{fig:trace_kde} shows kernel density estimates of the TRACE Stability Score across five math benchmarks, with solid curves corresponding to correct predictions and dashed curves to incorrect ones.

Across all datasets, correct predictions consistently concentrate in the high-score region (approximately \(0.8\) to \(1.0\)), indicating that successful reasoning is typically accompanied by stable answers and coherent confidence trajectories over multiple steps.
In contrast, incorrect predictions exhibit broader distributions and place substantially more probability mass at lower scores.
This suggests that reasoning failures are often associated with unstable intermediate answers or transient confidence patterns, which TRACE is designed to detect.

The degree of separation varies by dataset.
On benchmarks such as MATH and AMC23, incorrect predictions rarely achieve very high stability scores, resulting in a clear gap between correct and incorrect distributions.
For more challenging datasets, including AIME24, AIME25, and Olympiad Bench, the distributions partially overlap, reflecting cases where the model reaches a locally stable but ultimately incorrect conclusion.
Nevertheless, even in these settings, correct predictions remain strongly skewed toward higher stability values.

Overall, this analysis supports the use of a fixed termination threshold (e.g., \(\tau = 0.8\)) across datasets.
Most correct predictions lie above the threshold, while a substantial fraction of incorrect predictions fall below it, enabling TRACE to terminate reliably only after the reasoning process has genuinely converged.

\end{document}